\documentclass[10pt,twocolumn,letterpaper]{article}

\usepackage{cvpr}              

\usepackage{multirow}
\usepackage{pifont}
\usepackage{wrapfig}
\usepackage[ruled]{algorithm2e}
\usepackage{listings}
\usepackage{xcolor}

\definecolor{codegreen}{rgb}{0,0.6,0}
\definecolor{codegray}{rgb}{0.5,0.5,0.5}
\definecolor{codepurple}{rgb}{0.58,0,0.82}
\definecolor{backcolour}{rgb}{0.95,0.95,0.92}

\lstdefinestyle{mystyle}{
    backgroundcolor=\color{backcolour},   
    commentstyle=\color{codegreen},
    keywordstyle=\color{magenta},
    numberstyle=\tiny\color{codegray},
    stringstyle=\color{codepurple},
    basicstyle=\ttfamily\footnotesize,
    breakatwhitespace=false,         
    breaklines=true,                 
    captionpos=b,                    
    keepspaces=true,                 
    numbers=none,                    
    numbersep=5pt,                  
    showspaces=false,                
    showstringspaces=false,
    showtabs=false,                  
    tabsize=2
}
\makeatletter
\newcommand{\removelatexerror}{\let\@latex@error\@gobble}
\makeatother

\definecolor{cvprblue}{rgb}{0.21,0.49,0.74}
\usepackage[pagebackref,breaklinks,colorlinks,citecolor=cvprblue]{hyperref}

\def\paperID{6438} 
\def\confName{CVPR}
\def\confYear{2024}

\newcommand{\xmark}{\ding{55}}%

\DeclareMathSymbol{\mlq}{\mathord}{operators}{``}
\DeclareMathSymbol{\mrq}{\mathord}{operators}{`'}

\DeclareCaptionLabelFormat{andtable}{#1~#2  \&  \tablename~\thetable}

\newcommand\Tstrut{\rule{0pt}{2.6ex}}
\newcommand\STstrut{\rule{0pt}{2.2ex}}

\title{Confronting Ambiguity in 6D Object Pose Estimation \\via Score-Based Diffusion on SE(3)}

\author{Tsu-Ching Hsiao, \ Hao-Wei Chen, \ Hsuan-Kung Yang, and Chun-Yi Lee%
\smallskip
\\
Elsa Lab, National Tsing Hua University\\
{\tt\small \{joehsiao, jaroslaw1007, hellochick\}@gapp.nthu.edu.tw}\\
{\tt\small cylee@cs.nthu.edu.tw}
}

\begin{document}
\maketitle
\begin{abstract}
Addressing pose ambiguity in 6D object pose estimation from single RGB images presents a significant challenge, particularly due to object symmetries or occlusions. In response, we introduce a novel score-based diffusion method applied to the $SE(3)$ group, marking the first application of diffusion models to $SE(3)$ within the image domain, specifically tailored for pose estimation tasks. Extensive evaluations demonstrate the method's efficacy in handling pose ambiguity, mitigating perspective-induced ambiguity, and showcasing the robustness of our surrogate Stein score formulation on $SE(3)$. This formulation not only improves the convergence of denoising process but also enhances computational efficiency. Thus, we pioneer a promising strategy for 6D object pose estimation.
\end{abstract}    
\vspace{-1em}
\section{Introduction}
\label{sec:introduction}

Estimating the six degrees of freedom (DoF) pose of objects from a single RGB image remains a formidable task, primarily due to the presence of ambiguity induced by symmetric objects and occlusions.
Symmetric objects exhibit identical visual appearance from multiple viewpoints, whereas occlusions arise when key aspects of an object are concealed either by another object or its own structure. This can complicate the determination of its shape and orientation.
Pose ambiguity presents a unique challenge as it transforms the direct one-to-one correspondence between an image and its associated object pose into a complex one-to-many scenario, which can potentially leads to significant performance degradation for methods reliant on one-to-one correspondence.
Despite extensive exploration in the prior object pose estimation literature~\cite{manhardt2019,epos,dbn,implicitpdf,hodan2017tless}, pose ambiguity still remains a persisting and unresolved issue.

Recent advancements in pose regression have introduced the use of symmetry-aware annotations to improve pose estimation accuracy~\cite{manhardt2019,pix2pose,gdrnet,cope}. These methods typically employ symmetry-aware losses that can tackle the pose ambiguity problem.
The efficacy of these losses, nevertheless, depend on the provision of symmetry annotations, which can be particularly challenging to obtain for objects with intricate shapes or under occlusion.
An example is a texture-less cup, where the true orientation becomes ambiguous if the handle is not visible.The manual labor and time required to annotate the equivalent views of each object under such circumstances is impractical.

\begin{figure}[t]
  \centering
  \includegraphics[width=\linewidth]{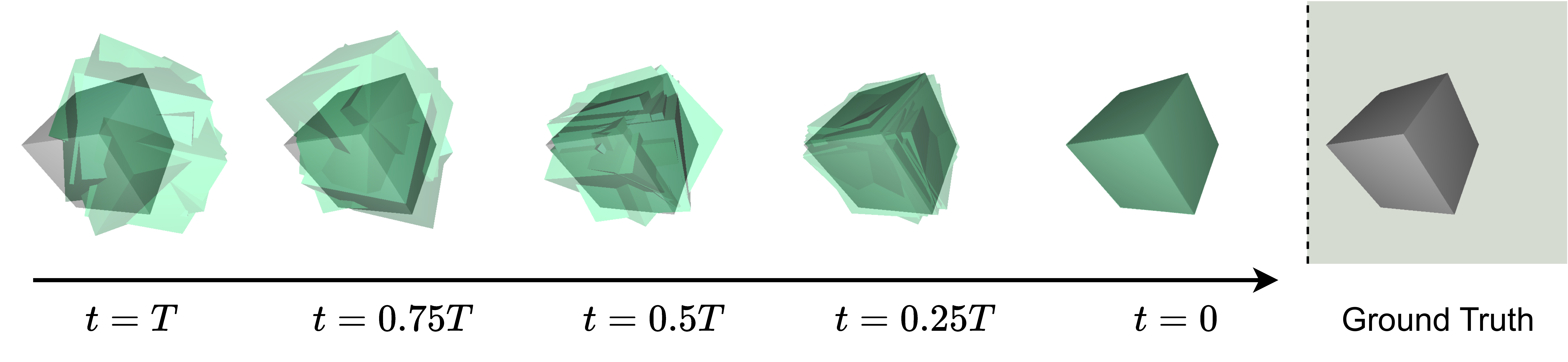}
  \caption{
  Visualization of the denoising process of our score-based diffusion method on $SE(3)$ for 6DoF pose estimation.}
  \label{fig:sample-process}
  \vspace{-1em}
\end{figure}

Several contemporary studies have sought to eliminate the reliance on symmetry annotations by treating `equivalent poses' as a multi-modal distribution, reframing the original pose estimation problem as a density estimation problem. Methods such as Implicit-PDF~\citep{implicitpdf} and HyperPose-PDF~\citep{hyperpose} leverage neural networks to implicitly characterize the non-parametric density on the rotation manifold $SO(3)$. While these advances are noteworthy, they also introduce new complexities. For instance, the computation during training requires exhaustive sampling across the whole $SO(3)$ space. Moreover, the accuracy of inference is dependent on the resolution of the grid search, which necessitates a significant amount of grid sampling. These computational limitations are magnified when extending to larger spaces such as $SE(3)$ due to the substantial memory requirements.

Recognizing these challenges, the research community is pivoting towards diffusion models (DMs)~\citep{ncsn, ddpm, ddim, scoresde}, which are effective in handling multi-modal distributions. Their effectiveness lies in the iterative sampling process, which incorporates noises and enables a more focus exploration of the pose space while reducing computational demands. As diffusion models refrain from explicit density estimation, this property enables them to handle large spaces and high-dimensional distributions. In prior endeavors, the authors in~\cite{leach2022, jagvaral2023} applied the denoising diffusion probabilistic model (DDPM)~\citep{ddpm} and score-based generative model~(SGM)~\citep{scoresde} to the $SO(3)$ rotation manifold, effectively recovering unknown densities on the $SO(3)$ space. On the other hand, other research efforts~\cite{urain2022, yim2023} have extended the application of diffusion models to the more complex $SE(3)$ space, which enlightens the potential applicability of diffusion models in object pose estimation tasks.

In light of the above motivations, we introduce a novel approach that applies diffusion models to the $SE(3)$ group for object pose estimation tasks, specifically aimed at addressing the pose ambiguity problem. This method draws its inspiration from the correlation observed between rotation and translation distributions, a phenomenon often resultant from the perspective effect inherent in image projection. We propose that by jointly estimating the distribution of rotation and translation on $SE(3)$, we may secure more accurate and reliable results as shown in Fig.~\ref{fig:sample-process}. To the best of our knowledge, this is the first work to apply diffusion models to $SE(3)$ within the context of image space. To substantiate our approach, we have developed a new synthetic dataset, called SYMSOL-T, based on the original SYMSOL dataset~\citep{implicitpdf}. It enhances the original dataset with randomly sampled translations, offering a more rigorous testbed to evaluate our method's effectiveness in capturing the joint density of object rotations and translations.

Following the motivations discussed above, we have extensively evaluated our $SE(3)$ diffusion model using the synthetic SYMSOL-T dataset and a real-world T-LESS~\citep{t-less} dataset. The experimental results affirm the model's competence in handling $SE(3)$, which successfully addresses the pose ambiguity problem in 6D object pose estimation. Moreover, the $SE(3)$ diffusion model has proven effective in enhancing rotation estimation accuracy and robustness. Importantly, the surrogate Stein score formulation we propose on $SE(3)$ exhibits improved convergence in the denoising process compared to the score calculated via automatic differentiation. This not only highlights the robustness of our method, but also demonstrates its potential to handle complex dynamics in object pose estimation tasks.
\section{Background}
\label{sec:background}

\subsection{Lie Groups and Their Applications}
\label{sec:background:lie_group}

A Lie group, denoted by $ \mathcal{G} $, serves as a mathematical structure with broad applicability due to its dual nature as both a group and a smooth (or differentiable) manifold. The latter is a topological space that can be locally approximated as a linear space. In accordance with the axioms governing groups, a composition operation is formally defined as a mapping $ \circ : \mathcal{G} \times \mathcal{G} \to \mathcal{G} $. The composition operation, along with the associated inversion map, exhibits smoothness properties consistent with the group structure. For notational convenience in subsequent analyses, the composition of two group elements $ X, Y \in \mathcal{G} $ is succinctly denoted as $ X \circ Y = XY $. Every Lie group $\mathcal{G}$ has an associated Lie algebra, denoted as $\mathfrak{g}$. A Lie group and its associated Lie algebra are related through the following mappings: $\text{Exp} : \mathfrak{g} \rightarrow \mathcal{G},~ \text{Log} : \mathcal{G} \rightarrow \mathfrak{g}$. In the context of pose estimation, two Lie groups are commonly employed: $ SO(3) $ and $ SE(3) $. The Lie group $ SO(3) $ and its associated Lie algebra $ \mathfrak{so}(3) $ can represent rotations in three-dimensional Euclidean space. On the other hand, the Lie group $ SE(3) $, along with its corresponding Lie algebra $ \mathfrak{se}(3) $, can be employed to describe rigid-body transformations, which incorporate both rotational and translational elements in Euclidean space.  Such group structures form the mathematical basis for analyzing and solving complex problems,  especially for six Degrees of Freedom (6DoF) pose estimation.

\subsection{Lie Group Representation of Transformations}
\label{sec:background:transformation}

A variety of parametrizations for these transformation groups are discussed in~\cite{sola2018}. This work considers two types of transformation groups, each characterized by a distinct manifold structure and the accompanying parametrizations: $R^3SO(3)$ and $SE(3)$. The former parametrization, which segregates rotations $R \in SO(3)$ and translations $T \in \mathbb{R}^3$ into a composite manifold $\langle\mathbb{R}^3,SO(3)\rangle$, denotes its Lie algebra as $\langle\mathbb{R}^3,\mathfrak{so}(3)\rangle$. $R^3SO(3)$ employs a composition rule defined by $(R_2, T_2)(R_1, T_1) = (R_2R_1, T_2 + T_1)$. This parametrization, which is prevalent in several prior diffusion models on $R^3SO(3)$ due to its simplicity as discussed in~\citep{yim2023,urain2022}, induces a separate diffusion process for both $R$ and $T$. Another parametrization, $SE(3)$, formulates elements within the Lie algebra as $\tau = (\rho, \phi) \in \mathfrak{se}(3)$, wherein $\rho$ and $\phi$ correspond to infinitesimal translations and rotations at the identity element's tangent space, respectively. The corresponding group elements within $SE(3)$ are represented as $(R, T) = (\text{Exp}(\phi), \mathbf{J}_l(\phi)\rho)$, where $\mathbf{J}_l$ denotes the left-Jacobian of $SO(3)$. The composition rule for the $SE(3)$ parametrization is expressed as $(R_2, T_2)(R_1,T_1) = (R_2R_1, T_2 + R_2T_1)$. The integration of both rotations and translations within $SE(3)$ gives rise to a diffusion process that emulates the elaborate dynamics of rigid-body motion.

\subsection{Score-Based Generative Modeling}
\label{sec:background:score_based}

Consider independent and identically distributed (i.i.d.) samples $\{ \mathbf{x}_{i} \in \mathbb{R}^{D} \}^{N}_{i=1}$ drawn from a data distribution $p_{\text{data}}(\mathbf{x})$. The \textit{(Stein) score} of a probability density $p(\mathbf{x})$ is the gradient of its logarithm, denoted as $\nabla_\mathbf{x}\log p(\mathbf{x})$~\citep{scorematching}. In the framework of score-based generative models (SGMs), an important formulation within the spectrum of diffusion models, data undergo a gradual transformation toward a known prior distribution. Such a distribution is often selected for computational tractability~\citep{vincent2011}, and this process is termed the \textit{forward} process. The forward process is characterized by a series of increasing noise levels $\{ \sigma_{i} \}^{L}_{i=1}$, which are ordered such that $\sigma_{\text{min}}=\sigma_{1} < \sigma_{2} < \ldots < \sigma_{L}=\sigma_{\text{max}}$. The selection of $\sigma_{\text{min}}$ and $\sigma_{\text{max}}$ as sufficiently small and large values respectively facilitates the approximation of $p_{\sigma_{\text{min}}}(\mathbf{x})$ to $p_{\text{data}}(\mathbf{x})$ and of $p_{\sigma_{\text{max}}}(\mathbf{x})$ to the Gaussian distribution $\mathcal{N}(\mathbf{x}; \mathbf{0}, \sigma_{\text{max}}^{2} \mathbf{I})$. This process utilizes a perturbation kernel $p_{\sigma}(\tilde{\mathbf{x}} | \mathbf{x}) = \mathcal{N}(\tilde{\mathbf{x}}; \mathbf{x}, \sigma^{2} \mathbf{I})$, and the perturbed distribution is given by $p_{\sigma}(\tilde{\mathbf{x}}) = \int p_{\text{data}}(\mathbf{x}) p_{\sigma}(\tilde{\mathbf{x}} | \mathbf{x}) d\mathbf{x}$. In the Noise Conditional Score Network (NCSN)~\citep{ncsn}, a network $s_{\boldsymbol{\theta}}(\mathbf{x}, \sigma)$ parameterized by $\theta$ is trained to estimate the score via a Denoising Score Matching (DSM) objective~\citep{vincent2011} as follows:
\vspace{-1em}
\begin{equation}
{\scriptsize
\begin{split}
    \boldsymbol{\theta}^{\ast} &= \mathop{\arg\min}_{\boldsymbol{\theta}} \mathcal{L}(\boldsymbol{{\theta}}; \sigma) \\
                              &\triangleq \frac{1}{2} \mathbb{E}_{p_{\text{data}}(\mathbf{x})}\mathbb{E}_{\tilde{\mathbf{x}} \sim \mathcal{N}(\mathbf{x}, \sigma^{2}I)} \left[ \left\| s_{\boldsymbol{\theta}}(\tilde{\mathbf{x}}, \sigma) - \nabla_{\tilde{\mathbf{x}}} \log p_{\sigma}(\tilde{\mathbf{x}} | \mathbf{x}) \right\|^{2}_{2} \right].
\end{split}
}
\label{eq:score_matching}
\end{equation}

The optimal score-based model $s_{\boldsymbol{\theta}^{\ast}}(\mathbf{x}, \sigma)$ aims to match $\nabla_{\mathbf{x}} \log p(\mathbf{x})$ as closely as possible across the entire range of $\sigma$ values in the set $\{\sigma_{i}\}^{L}_{i=1}$. During the sample generation phase, score-based generative models employ an iterative \textit{reverse} process. Specifically, in the context of the Noise Conditional Score Network (NCSN), the Langevin Markov Chain Monte Carlo (MCMC) method is utilized to execute $M$ steps. This process is designed to produce samples in a sequential manner from each $p_{\sigma_{i}}(\mathbf{x})$, expressed as follows:
\vspace{-1em}
\begin{equation}
\footnotesize
    \tilde{\mathbf{x}}^{m}_{i} = \tilde{\mathbf{x}}^{m-1}_{i} + \epsilon_{i} s_{\boldsymbol{\theta}^{\ast}}(\tilde{\mathbf{x}}^{m-1}_{i}, \sigma_{i}) + \sqrt{2\epsilon_{i}} \mathbf{z}^{m}_{i}, \quad m=1, 2, ..., M,
    \label{eq:langevin_dynamics}
\end{equation}
where $\epsilon_{i} > 0$ denotes the step size, and $\mathbf{z}^{m}_{i}$ represents a standard normal variable. Overall, diffusion based models, especially SGMs, provide a solid framework for handling complex data distributions. They serve as the foundation for the denoising procedure employed by our methodology.

\section{Related Work}

\begin{table*}
	\centering
	\caption{Comparison of different methods. $\triangle$ means closed form but with approximation. $\mathcal{N}_{SE(3)}$ please refer to Eq.~(\ref{eq:perturbation_lie}).}
    \vspace{-1em}
	\resizebox{.8\linewidth}{!}{
	\begin{tabular}{c|c|c|c|c|c|c}
    \specialrule{0.8pt}{0.0ex}{0.2ex}
    \textbf{Baselines} & \textbf{Group} & \textbf{Distribution} & \textbf{Closed Form} & \textbf{Diffusion Method} & \textbf{Diffusion Space} & \textbf{App. Domain}\\ \hline
    \STstrut Leach \etal~\cite{leach2022} & $SO(3)$ & $IG_{SO(3)}$ & \xmark & DDPM & $SO(3)$ & Vector \\ \hline
    \Tstrut Jagvaral \etal~\cite{jagvaral2023} & $SO(3)$ & $IG_{SO(3)}$ & \xmark & Score / Autograd & $SO(3)$ & Vector \\ \hline
    \Tstrut Urain \etal~\citep{urain2022} & $R^3SO(3)$ & $\mathcal{N}_{\mathbb{R}^3} \times \mathcal{N}_{SO(3)}$ & \checkmark & Score / Autograd & $R^3SO(3)$ & Vector \\ \hline
    \Tstrut Yim \etal~\citep{yim2023} & $R^3SO(3)$ & $\mathcal{N}_{\mathbb{R}^3} \times IG_{SO(3)}$ & \xmark & Score / Autograd & $\langle\mathbb{R}^3,\mathfrak{so}(3)\rangle$ & Vector \\ \hline
    \STstrut Ours & $SE(3)$ & $\mathcal{N}_{SE(3)}$ & $\triangle$ & Score / Closed Form & $SE(3)$  & Image \\ 
    \specialrule{0.8pt}{0.2ex}{0.0ex}
    \end{tabular}}
    \vspace{-1em}
    \label{tab:comparison_and_ambiguity}
\end{table*}

\subsection{Methodologies for Dealing with Pose Ambiguity}

\paragraph{Non-probabilistic modeling.}~In the realm of object pose estimation, pose ambiguity remains a significant challenge, often stemming from an object that exhibits identical visual appearances from different perspectives~\cite{manhardt2019}. A variety of strategies have been explored in the literature to directly address this issue, including the application of symmetry supervisions and point matching algorithms~\cite{posecnn,yolopose}. Regression-based approaches, such as those presented in~\cite{labbe2020cosypose,gdrnet,sopose,cope}, aim to minimize pose discrepancy by selecting the closest candidate within a set of ambiguous poses. Some researchers~\cite{pvnet1,bb8}, on the other hand, introduce constraints to the regression targets (especially regarding rotation angles) to mitigate ambiguity.  Moreover,  certain approaches~\cite{pix2pose,wang2019normalized,huang2022} suggest regressing to a predetermined set of geometric features derived from symmetry annotations. These prior arts often necessitate manual annotations of equivalent poses and are limited in dealing with other sources of pose ambiguities, such as those caused by occlusion and self-occlusion~\cite{manhardt2019}.

\paragraph{Probabilistic modeling.}~On the other hand, several studies have investigated methods to model the inherent uncertainty in pose ambiguity. This involves the quantification and representation of uncertainty associated with the estimated poses. Some works have employed parametric distributions such as Bingham distributions~\cite{okorn2020learning,gilitschenski2020deep,dbn} and von-Mises distributions~\cite{prokudin2018deep,yin2023a} to model orientation uncertainty. Other approaches, such as in~\cite{liu2023delving}, utilize normalizing flows~\cite{rezende2020normalizing} to model distributions within rotational space. A number of studies~\cite{implicitpdf,hyperpose,i2s} employ non-parametric distributions to implicitly represent rotation uncertainty on $SO(3)$. These methods primarily focus on modeling distributions on $SO(3)$, leaving the joint distribution modeling of rotation and translation unexplored.

\subsection{Diffusion Probabilistic Models and Their Application Domains}

\paragraph{Diffusion models on Euclidean space.}~Diffusion probabilistic models~\cite{yang2022diffusionsurvey,ddpm, ddim, scoresde, ncsn} represent a class of generative models designed to learn the underlying probability distribution of data. They have been applied to various generative tasks, and have shown impressive results in several application domains, including image~\cite{ramesh2022hierarchical,ruiz2022dreambooth,rombach2022high,saharia2022photorealistic,amit2021segdiff,baranchuk2021label,diffusiondet}, video~\cite{yang2022diffusion,ho2022video,ho2022imagen}, audio~\cite{huang2022prodiff,yang2023diffsound}, and natural language processing~\cite{gong2022diffuseq,li2022diffusion}.
In the realm of human pose estimation, diffusion models have also been found useful in addressing joint location ambiguity, which arises from the projection of 2D keypoints into 3D space~\cite{diffupose,holmquist2022diffpose}.

\paragraph{Diffusion models on non-Euclidean space.}~To accommodate data residing on a manifold, the authors in~\cite{bortoli2022riemannian} extended diffusion models to Riemannian manifolds, and leveraged Geodesic Random Walk~\cite{jorgensen1975central} for sampling. Other studies~\cite{jagvaral2023, leach2022} applied the Denoising Diffusion Probabilistic Models (DDPM)~\cite{ddpm} and score-based generative models~\cite{scoresde, ncsn} to the $SO(3)$ manifold to recover the density of data on $SO(3)$. Further extensions of diffusion models have been attempted for tasks such as unfolding protein structures~\cite{yim2023} and arm manipulations~\cite{urain2022}. These approaches typically used $R^3SO(3)$ parametrization, which treated rotation and translation as separate entities for diffusion.

\subsection{Diffusion Models on Lie Groups}
\label{subsec::diffusion_lie_group}

Diffusion models on Lie groups have been explored in a range of applications~\citep{leach2022, jagvaral2023, urain2022, yim2023}. Nevertheless, these implementations vary in their choices of distributions and computational methods, which lead to diverse outcomes and different levels of computational efficiency. Table~\ref{tab:comparison_and_ambiguity} presents a comparison of several previous diffusion model approaches along with our own. It highlights the distinct groups, distributions, methods, as well as diffusion spaces each method utilizes. Several earlier studies~\cite{leach2022, jagvaral2023} have introduced techniques that operate within the $SO(3)$ space, and adopted normal distributions defined on $SO(3)$~\citep{nikolayev1970normal} (denoted as $IG_{SO(3)}$). Unfortunately, a primary drawback of $IG_{SO(3)}$ is its absence of a closed form, which poses challenges in its computational efficiency. In a similar vein, the authors in~\cite{yim2023} developed a method that operates in the tangent space of $R^3SO(3)$. This method's distribution also does not possess a closed form, which complicates the computational procedure. On the other hand, the authors in~\cite{urain2022} employed a joint Gaussian distribution within the $\mathbb{R}^3$ and $SO(3)$ spaces. This distribution benefits from the presence of a closed form and thus offers the potential for increased computational efficiency. However, this approach is confined to the $\mathbb{R}^3 \times SO(3)$ space and treats rotation and translation as separate entities for diffusion. As a result, it may not be able to offer the advantages that  $SE(3)$ can provide.

\section{Methodology}
\label{sec:methodology}

Given an RGB image $I$ that displays the object of interest, our goal is to estimate the 6D object poses $X=(R,T)\in SE(3)$, which represent the transformation from the camera frame to the object. This estimation involves sampling poses from a conditional distribution $X\sim p(X|I)$, which captures the inherent pose uncertainty of the object depict in $I$. To facilitate this process, our method employs a score-based generative model on $SE(3)$ to recover this underlying distribution. Poses are then sampled via a \textit{reverse} process that gradually refines noisy pose hypotheses $\tilde{X}\sim p(\tilde{X})$ drawn from a known prior distribution $p(\tilde{X})$, specifically a Gaussian distribution on $SE(3)$. Both the \textit{forward} and \textit{reverse} processes are performed on Lie groups and leverage the associated group operations. It is important to note that our approach does not utilize 3D models of the objects or symmetry annotations during either the training or inference phases, instead relying exclusively on RGB images and the associated ground truth (GT) poses for training.

\subsection{Score-Based Pose Diffusion on a Lie Group}
\label{sec:methodology:diffusion_framework}

\newcommand{\Log}{\text{Log}}
\newcommand{\Exp}{\text{Exp}}
\newcommand{\gauss}{\mathcal{N}}
\newcommand{\group}[1]{\mathcal{#1}}
\newcommand{\tang}[1]{\mathfrak{#1}}
\newcommand{\vecbf}[1]{\mathbf{#1}}

To apply score-based generative modeling to a Lie group $\mathcal{G}$, we first establish a perturbation kernel on $\mathcal{G}$ that conforms to the Gaussian distribution~\cite{riemannian,chirikjian2014gaussian}. The kernel is given by:
\vspace{-1em}
\begin{equation}
{\scriptsize
\begin{split}
p_\Sigma(Y|X) &:= \mathcal{N}_{\mathcal{G}}(Y;X,\Sigma) \\
              &\triangleq \frac{1}{\zeta(\Sigma)}\exp\left(-\frac{1}{2}\Log(X^{-1}Y)^\top\Sigma^{-1}\Log(X^{-1}Y)\right),
\end{split}
}
\label{eq:perturbation_lie}
\end{equation}
where $\Sigma$ is the covariance matrix with diagonal entries populated by $\sigma$ for representing the scale of the perturbation, $\zeta(\Sigma)$ is the normalizing constant, and $X, Y\in \mathcal{G}$ denote the group elements. The \textit{score} on $\mathcal{G}$ then corresponds to the gradient of the log-density of the data distribution with respect to the group element $Y$. It can be formulated as follows:
\begin{equation}
\footnotesize
\nabla_Y \log p_\Sigma(Y|X) = -\mathbf{J}_r^{-\top}(\Log(X^{-1}Y))\Sigma^{-1}\Log(X^{-1}Y).
\label{eq:score_form}
\end{equation}
This term can be expressed in closed form if the inverse of the right-Jacobian $\mathbf{J}_r^{-1}$ on $\mathcal{G}$ exists in a closed form. Nevertheless, an alternative approach suggested by the authors in~\cite{urain2022} would be to compute this term using automatic differentiation~\citep{paszke2017automatic}. By substituting $Y$ with $\tilde{X}$, assuming $\tilde{X} = X \Exp(z), ~z \sim \mathcal{N}(0, \sigma_i^2I)$, and integrating the above definition, the \textit{score} on $\mathcal{G}$ can be reformulated as follows:
\vspace{-0.5em}
\begin{equation}
\nabla_{\tilde{X}} \log p_\sigma(\tilde{X}|X) = -\frac{1}{\sigma^2}\mathbf{J}_r^{-\top}(z)z.
\label{eq:score_jacob}
\end{equation}
A score model $s_{\boldsymbol{\theta}}(\tilde{X}, \sigma)$ can then be trained using the DSM objective shown in Eq.~(\ref{eq:score_matching}), which takes the following form:
\vspace{-1em}
\begin{equation}
{\scriptsize
\begin{split}
     \boldsymbol{\theta}^{\ast} &= \mathop{\arg\min}_{\boldsymbol{\theta}} \mathcal{L}(\boldsymbol{{\theta}}; \sigma) \\
                                \triangleq &\frac{1}{2} \mathbb{E}_{p_{\text{data}}(X)}\mathbb{E}_{\tilde{X} \sim \mathcal{N}_{\mathcal{G}}(X, \Sigma)} \left[ \left\| s_{\boldsymbol{\theta}}(\tilde{X}, \sigma) - \nabla_{\tilde{X}} \log p_{\sigma}(\tilde{X} | X) \right\|^{2}_{2} \right].  
\end{split}
}
\label{eq:score_matching_lie}
\end{equation}
For the denoising process, we employ a variant of the Geodesic Random Walk~\citep{bortoli2022riemannian}, tailored to the Lie group context, as a means to generate a sample from a noise distribution. The procedure is expressed as follows:
\begin{equation}
{\small
\begin{split}
    \tilde{X}_{i+1} = \tilde{X}_{i}\Exp(\epsilon_i s_\theta(\tilde{X}_{i}, \sigma_i) + \sqrt{2\epsilon_i} z_i), \quad z_i\sim \mathcal{N}(0, I).
\end{split}
}
\label{eq:langevin_dynamics_lie}
\end{equation}

\subsection{Efficient Computation of the Stein Score}
\label{sec:methodology:simplify_score}
Even with the above derivation, obtaining the closed-form \textit{score} remains challenging due to its dependency on the selected distribution. For instance, deriving the closed-form \textit{score} for the $IG_{SO(3)}$ distribution~\cite{nikolayev1970normal} poses difficulties. Furthermore, computing the \textit{score} depends on the existence of a closed-form expression for the Jacobian matrix on $\mathcal{G}$. Even if such an expression exists, it may not guarantee computational efficiency compared to automatic differentiation. Therefore, we next discuss a simplification method of the Stein \textit{score} under certain conditions for reducing computational costs on $\mathcal{G}$. This can be expressed in a closed-form if the Jacobian matrix on $\mathcal{G}$ is invertible and if the left and right Jacobian matrices conform to the following relation:
\begin{equation}
\mathbf{J}_l(z) = \mathbf{J}^\top_r(z),\quad \mathbf{J}_l^{-1}(z) = \mathbf{J}_r^{-\top}(z),
\label{eq:left_right_jacob}
\end{equation}
where $z \in \mathfrak{g}$. As pointed out by~\cite{sola2018}, $SO(3)$ exhibits this property. Its closed-form \textit{score} can then be simplified by utilizing the following property, which holds on any $\mathcal{G}$ as $\mathbf{J}_l(z)z = z$.
The derivation is in the supplementary material. The \textit{score} on $SO(3)$ can then be expressed as follows:
\begin{equation}
\nabla_{\tilde{X}}\log p_\sigma(\tilde{X}|X) = -\frac{1}{\sigma^2} \mathbf{J}_{l}^{-1}(z)z = -\frac{1}{\sigma^2} z.
\label{eq:score_simplified}
\end{equation}
This shows that the \textit{score} on $SO(3)$ can be simplified to the sampled Gaussian noise $z$ scaled by $-1/{\sigma^{2}}$, thus eliminating the need for both automatic differentiation and Jacobian calculations. Similarly, the \textit{score} on $R^3SO(3)$ also has a closed-form as its Jacobians satisfy the relations in Eq.~(\ref{eq:left_right_jacob}):
\begin{equation}
    \mathbf{J}_l(z) = (I, \mathbf{J}_l(\phi)) = (I, \mathbf{J}_r^\top(\phi)) = \mathbf{J}_r^\top(z),
\end{equation}
where in the case of $R^3SO(3)$, $z=(T, \phi)\in\langle\mathbb{R}^3,\mathfrak{so}(3)\rangle$. This implies that the \textit{score} on $R^3SO(3)$ can also be simplified according to the formulation represented by Eq.~(\ref{eq:score_simplified}).

\begin{figure*}
    \centering
    \resizebox{.9\linewidth}{!}{%
    \begin{minipage}[t]{.672\textwidth}
        \centering
        \includegraphics[width=\textwidth]{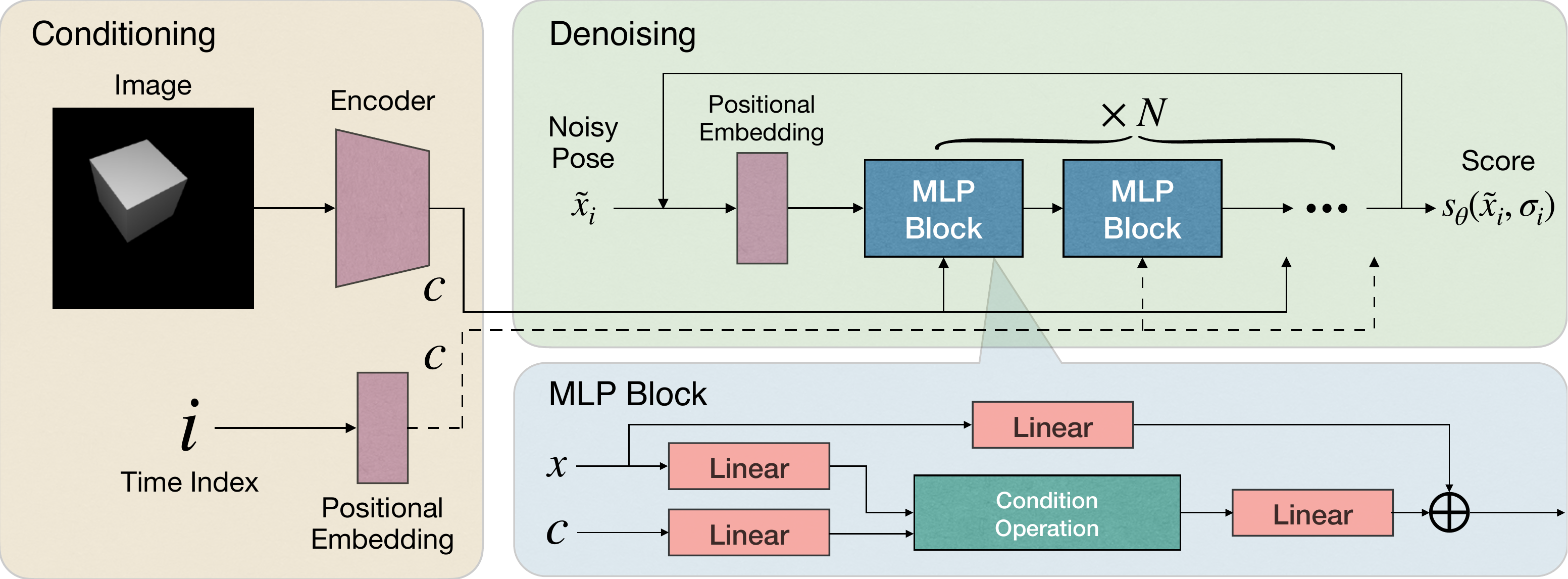}
        \phantomsubcaption \label{subfiglcr:left}
    \end{minipage}
    \hfill
    \begin{minipage}[t]{.31\textwidth}
        \centering
        \includegraphics[width=\textwidth]{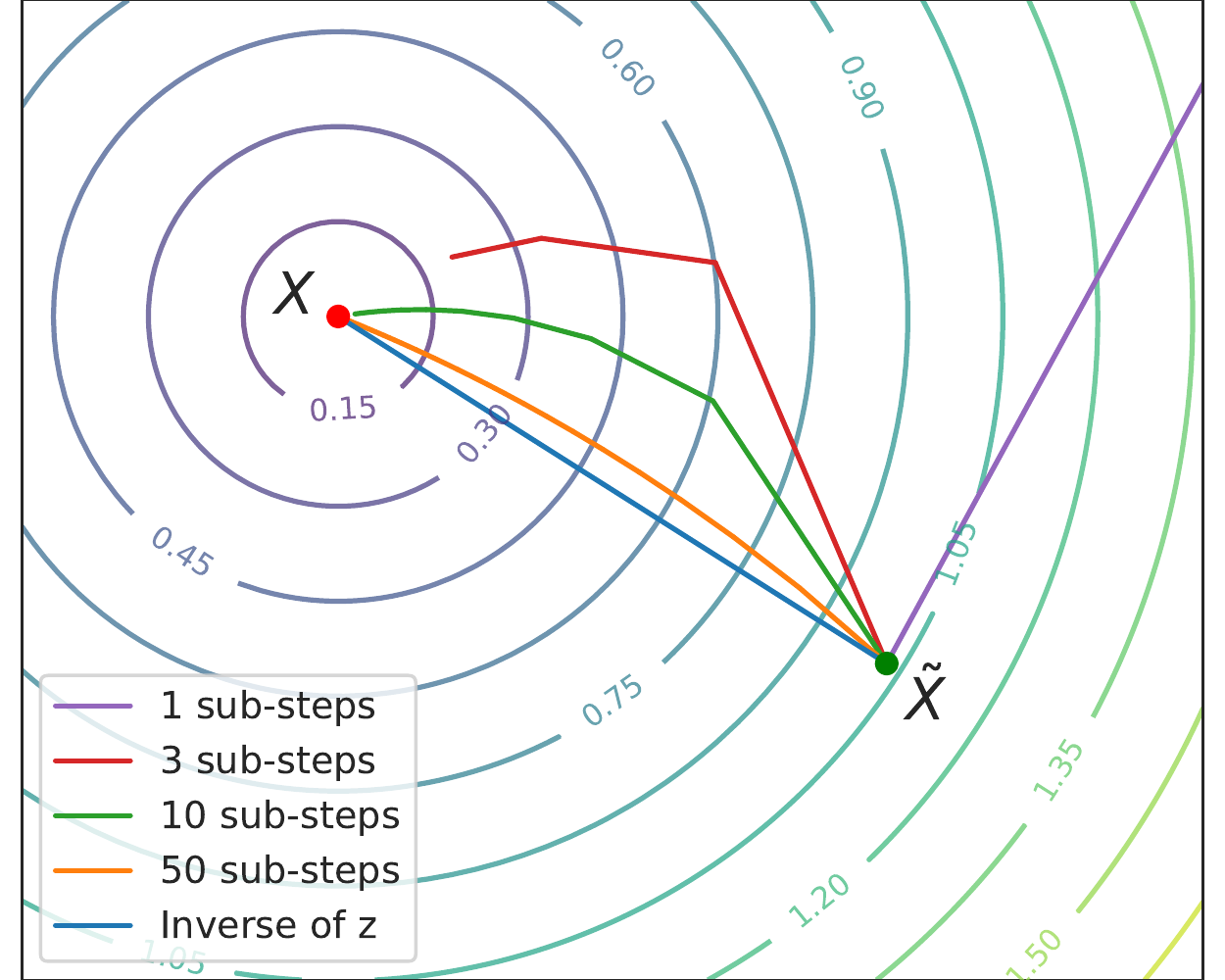}
        \phantomsubcaption \label{subfiglcr:right}
    \end{minipage}
    }
    \vspace{-1em}
    \caption{\textbf{Left:} Framework overview. \textbf{Right:} Visualization of a denoising step from a noisy sample $\tilde{X}$ to its cleaned counterpart $X$ on $SE(2)$. The contours are the distances to $X$ in 2D Euclidean space. Each line represents a denoising path with varying sub-sampling steps.}
    \label{fig:all}
    \vspace{-1em}
\end{figure*}

\subsection{Surrogate Stein Score Calculation on $SE(3)$}
\label{sec:methodology:simplify_score_se3}

While the \textit{score} on $SO(3)$ and $R^3SO(3)$ can be simplified as described in the preceding sections, it can be shown that $SE(3)$ does not possess the property in Eq.~(\ref{eq:left_right_jacob}).
Consider the inverse of the left-Jacobian on $SE(3)$ at $z=(\rho, \phi)\in\mathfrak{se}(3)$, expressed as $\mathbf{J}_l^{-1}(z)=\left[\begin{smallmatrix}\mathbf{J}_l^{-1}(\phi) & \mathbf{Z}(\rho, \phi)\\0 & \mathbf{J}_l^{-1}(\phi)\\\end{smallmatrix}\right]$, where $\mathbf{Z}(\rho, \phi)=-\mathbf{J}^{-1}_l(\phi)\mathbf{Q}(\rho, \phi)\mathbf{J}^{-1}_l(\phi)$. The complete form of $\mathbf{Q}(\rho, \phi)$ can be found in~\citep{sola2018, barfoot2014} and our supplementary material. The property $\mathbf{Q}^\top(-\rho, -\phi) = \mathbf{Q}(\rho, \phi)$, as derived in the references, leads to the following inequality:
\begin{equation}
{\footnotesize
\begin{split}
    \mathbf{J}_r^{-\top}(z) = (\mathbf{J}_l^{-1}(-z))^\top = \left[\begin{smallmatrix} \mathbf{J}^{-1}_l(\phi) & 0\\ \mathbf{Z}(\rho, \phi) & \mathbf{J}^{-1}_l(\phi) \end{smallmatrix}\right] \ne \mathbf{J}^{-1}_l(z).
\end{split}
}
\end{equation}
This inequality indicates the potential discrepancy between the \textit{score} vector and the denoising direction due to the curvature of the manifold, which may impede the convergence of the reverse process and necessitate additional denoising steps.
To address this problem, we turn to higher-order approximation methods by breaking one step of reverse process into multiple smaller sub-steps. Fig.~\ref{fig:all}~(right) illustrates this one-step denoising process on $SE(2)$ from a noisy sample $\tilde{X}=X\text{Exp}(z)$ to its cleaned counterpart $X$, with contour lines representing the distance to $X$ in 2D Euclidean space. We observe that increasing the number of sub-steps eventually leads the integral of those \textit{small} transformations approaches the inverse  of $z$. As a result, we propose substituting the \textit{true score} in Eq.~(\ref{eq:score_jacob}) with a \textit{surrogate score} in our training objective of Eq.~(\ref{eq:score_matching_lie}) on $SE(3)$, defined as follows:
\begin{equation}
{\small
\begin{split}
\tilde{s}_X(\tilde{X}, \sigma) \triangleq -\frac{1}{\sigma^2} z.
\end{split}
}
\label{eq:surrogate_score}
\end{equation}
Note that the detailed training and sampling procedures are described and elaborated in our supplementary material.

\subsection{The Proposed Framework}
\label{sec:methodology:proposed_framework}

Fig.~\ref{fig:all}~(left) presents an overview of our framework, which consists of a conditioning part and a denoising part. The conditioning part is responsible for generating the condition variable $c$, which is crucial for guiding the denoising process. This variable $c$ can be derived either from an image encoder which extracts features from an image, or from a positional embedding module~\citep{vaswani2017attention} that encodes a time index $i$. In our experiments, we employ ResNet~\cite{he2016deep} as the image encoder. The separation of the two parts in our framework eliminates the need of image feature extraction in every denoising step, which offers efficiency in the inference phase. For the denoising part, our score model is composed of multiple multi-layer perceptron (MLP) blocks. This structure is inspired by the recent conditional generative models~\cite{ddpm, ncsn}, while we have modified their approaches by substituting linear layers for the convolutional ones. The score model processes a noisy pose $\tilde{x}_i\in\mathfrak{g}$ embedded using a positional encoding. It then computes an estimated \textit{score} $s_\theta(\tilde{x}_i, \sigma_i)$. This estimated \textit{score} is subsequently utilized in the denoising process (i.e., Eq.~(\ref{eq:langevin_dynamics_lie})).  Please note that the input and output of the denoising part are represented in vector forms within the corresponding Lie algebra space.

Regarding the design of the conditioning mechanism in MLPs, a few prior studies~\citep{ddpm,ncsn} employ scale-bias condition, which is  formulated as $f(x, c)=\mathbf{A}(c)x+\mathbf{B}(c)$. Nevertheless, our empirical observations suggest that this conditioning mechanism does not perform satisfactorily when learning distributions on $SO(3)$. This may be attributable to the limited expressivity of the underlying neural networks. Inspired by~\citep{ziyin2020neural,lee2021conditional}, we introduce a modified Fourier-based conditioning mechanism, which is formulated as follows:
\begin{equation}
{\footnotesize
\begin{split}
    f_i(x, c) = \sum^{d-1}_{j=0} \mathbf{W}_{ij}\left(\mathbf{A}_j(c)\cos(\pi x_j) + \mathbf{B}_j(c)\sin(\pi x_j)\right),
\end{split}
}
\end{equation}
where $d$ represents the dimension of our linear layer. This form bears similarity to the Fourier series $f(t) = \sum^\infty_{k=0} \mathbf{A}_k\cos\left(\frac{2\pi kt}{P}\right) + \mathbf{B}_k \sin\left(\frac{2\pi k t}{P}\right)$. Our motivation stems from the fact that the pose distribution on SO(3) is circular, and can therefore be represented as periodic functions. By the definition of periodic functions, their derivatives are also periodic. It is worth noting that this conditioning mechanism does not introduce additional parameters in our neural network design, as $\mathbf{W}_{ij}$ is provided by the subsequent linear layer. Our experimental findings suggest that this conditioning scheme enhances the ability of neural network to capture periodic features of score fields on $SO(3)$.

\section{Experimental Results}
\label{sec:experimental_results}

In this section, we demonstrate that our score-based diffusion model can produce precise pose estimation on both $SO(3)$ and $SE(3)$ compared with previous probabilistic approaches. In addition, we present our method's superior performance on the real-world T-LESS~\citep{t-less} dataset without relying on reconstructed 3D models or symmetric annotations. Note that, to the best of our knowledge, our approach is the first probabilistic model that conduct the experiments on the complete T-LESS dataset and reports the accuracy, in contrast to previous methods confined to a limited subset of objects. The extensive evaluation substantiate the robustness and scalability of our score-based diffusion model.

\subsection{Experimental Setups}
\label{sec:experimental_setups}

\paragraph{SYMSOL.}~SYMSOL is a dataset specifically designed for evaluating density estimators in the $SO(3)$ space. 
This dataset, first introduced by~\citep{implicitpdf}, comprises 250k images of five texture-less and symmetric objects, with each subject to random rotations. The objects include tetrahedron (tet.), cube, icosahedron (icosa.), cone, and cylinder (cyl.), with each exhibiting unique symmetries that introduce various degrees of pose ambiguity. For this dataset, our score model is compared in the $SO(3)$ space with several recent works~\citep{dbn, implicitpdf, hyperpose, Liu_2023_CVPR}. The baseline models compared with utilize a pre-trained ResNet50~\citep{resnet} as their backbones. Note that we report the average angular distances in degrees.

\paragraph{SYMSOL-T.} To extend our evaluation into the $SE(3)$ space, we developed the SYMSOL-T dataset by incorporating random translations based on SYMSOL, which introduces an additional layer of complexity due to perspective-induced ambiguity.
Similar to SYMSOL, it features the same five symmetric shapes and the same number of random samples.  For SYMSOL-T, we benchmark our proposed methods against two pose regression methods. These two methods are trained using a symmetry-aware loss, but with different strategies: one directly estimates the pose from an image, while the other employs iterative refinement. We report the average angular distances in degrees for rotation and the average distances for translation.

\paragraph{T-LESS.} T-LESS~\citep{t-less} has been recognized as a challenging benchmark in the BOP challenge~\citep{bop-challenge}, which consists of thirty texture-less industrial objects. The objects in this dataset are characterized by a range of discrete and continuous symmetries. In this dataset,  the pose ambiguities arise not only from the intrinsic object symmetries but also the environmental factors such as occlusion and self-occlusion due to its cluttered settings. The T-LESS dataset features a training set with 50k physically based rendering (PBR)~\citep{bop-challenge} images from synthetic images, and an additional 37k images from real-world scanning. The testing set encompasses 10k real-world scanned images. The evaluation methods employed in our study include three standard metrics from the BOP challenge: Maximum Symmetry-Aware Projection Distance (MSPD), Maximum Symmetry-Aware Surface Distance (MSSD), and Visible Surface Discrepancy (VSD). To reflect the emphasis of our work on symmetry, we further introduced symmetry-aware metrics: R@2, R@5, and R@10, which represent predictions with rotational errors within 2, 5, and 10 degrees, respectively.  Similarly, T@2, T@5, and T@10 are estimations with translational errors within 2, 5, and 10 centimeters, respectively.

\paragraph{Visualization} To visualize the density predictions, we adopt the strategy employed in~\cite{implicitpdf} to represent the rotation densities generated by our model in the $SO(3)$ space. Specifically, we use the Mollweide projection for visualizing the $SO(3)$ space, with longitude and latitude values representing the yaw and pitch of the object's rotation, respectively. The color in the $SO(3)$ space indicates the roll of the object's rotation. The circles denote sets of equivalent poses, with each dot representing a single sample. For each plot, we generate a total of $1,000$ random samples from our model. For the translation part, we illustrate the rendered results of the estimated poses below their original images.

\begin{table}[t]
    \centering
    \footnotesize	
    
    \centering
    \caption{Evaluation results on SYMSOL.}
    \vspace{-1em}
    \resizebox{\linewidth}{!}{%
    \begin{tabular}{l|c|c|c|c|c|c}
    \specialrule{0.8pt}{0.0ex}{0.2ex}
    \multirow{2}{*}{\textbf{Methods}} & \multicolumn{6}{c}{\textbf{SYMSOL (Spread in degrees~$\downarrow$)}} \\ \cline{2-7}
     & \textbf{Avg.} & \textbf{tet.} & \textbf{cube} & \textbf{icosa.} & \textbf{cone} & \textbf{cyl.} \\ 
    \hline
    DBN~\citep{dbn} & 22.44 & 16.70 & 40.70 & 29.50 & 10.10 & 15.20 \\ 
    Implicit-PDF~\citep{implicitpdf} & 3.96 & 4.60 & 4.00 & 8.40 & 1.40 & 1.40 \\ 
    HyperPosePDF~\citep{hyperpose} & 1.94 & 3.27 & 2.18 & 3.24 & 0.55 & 0.48 \\
    Normalizing Flows~\citep{Liu_2023_CVPR} & 0.70 & 0.60 & 0.60 & 1.10 & 0.50 & 0.50 \\ 
    \hline
    Ours (ResNet34) & 0.42 & 0.43 & 0.44 & 0.52 & \textbf{0.35} & 0.35 \\ 
    Ours (ResNet50) & \textbf{0.37} & \textbf{0.28} & \textbf{0.32} & \textbf{0.40} & 0.53 & \textbf{0.31} \\
    \specialrule{0.8pt}{0.2ex}{0.0ex}
    \end{tabular}}
    \label{tab:symsol_result}
    \vspace{-1em}
\end{table}

\begin{table}[t]
    \centering
    \caption{Evaluation results on SYMSOL-T.}
    \vspace{-1em}
    \resizebox{\linewidth}{!}{%
    \begin{tabular}{l|cc|cc|cc|cc|cc}
    \specialrule{0.8pt}{0.0ex}{0.2ex}
    \multirow{3}{*}{\textbf{Methods}} & \multicolumn{10}{c}{\STstrut \textbf{SYMSOL-T (Spread in degrees~$\downarrow$)}} \\ \cline{2-11}
    \STstrut & \multicolumn{2}{c|}{\textbf{tet.}} & \multicolumn{2}{c|}{\textbf{cube}} & \multicolumn{2}{c|}{\textbf{icosa.}} & \multicolumn{2}{c|}{\textbf{cone}} & \multicolumn{2}{c}{\textbf{cyl.}}\\ \cline{2-11}
    \STstrut & $R$ & $t$ & $R$ & $t$ & $R$ & $t$ & $R$ & $t$ & $R$ & $t$ \\ \hline
    \STstrut Regression & 2.92 & 0.064 & 2.86 & 0.05 & 2.46 & 0.037 & 1.84 & 0.058 & 2.24 & 0.049\\ 
    \STstrut Iterative regression & 4.25 & 0.048 & 4.2 & 0.037 & 29.33 & 0.026 & 1.63 & 0.037 & 2.34 & 0.032 \\ \hline
    \STstrut Ours ($R^{3}SO(3)$) & 1.38 & 0.017 & 1.93 & \textbf{0.010} & 29.35 & \textbf{0.009} & 1.33 & \textbf{0.016} & 0.86 & \textbf{0.010} \\ 
    \STstrut Ours ($SE(3)$) &  \textbf{0.59} & \textbf{0.016} & \textbf{0.58} & 0.011 & \textbf{0.64} & 0.012 & \textbf{0.54} & 0.016 & \textbf{0.41} & 0.011 \\
    \specialrule{0.8pt}{0.2ex}{0.0ex}
    \end{tabular}}
    \label{tab:symsolt_result}
    \vspace{-1em}
\end{table}

\begin{table}[t]
    \centering
    \caption{Evaluation results on T-LESS (Average of 30 objects).}
    \vspace{-1em}
    \resizebox{\linewidth}{!}{%
    \begin{tabular}{l|c|c|c||c|c|c||c|c|c}
    \specialrule{0.8pt}{0.0ex}{0.2ex}
    \multirow{2}{*}{\textbf{Methods}} & \multicolumn{9}{c}{\textbf{T-LESS (Accuracy~\% $\uparrow$)}} \\ \cline{2-10}
    & \textbf{MSPD} & \textbf{MSSD} & \textbf{VSD} & \textbf{R@2} & \textbf{R@5} & \textbf{R@10} & \textbf{T@2} & \textbf{T@5} & \textbf{T@10} \\
    \hline
    \STstrut GDRNPP~\citep{gdrnet} & 90.17 & 75.06 & 67.60 & 21.60 & 71.18 & 90.56 & 90.31 & 96.09 & 98.10 \\
    \hline
    \STstrut Ours ($R^{3}SO(3)$) & 85.73 & 52.03 & 48.41 & 27.98 & 72.42 & 89.26 & 60.37 & 79.75 & 89.62  \\ 
    \STstrut Ours ($SE(3)$) & \textbf{93.16} & 60.17 & 56.88 & \textbf{47.21} & \textbf{86.94} & \textbf{94.78} & 71.72 & 92.03 & 97.15 \\
    \specialrule{0.8pt}{0.2ex}{0.0ex}
    \end{tabular}}
    \label{tab:tless}
    \vspace{-1em}
\end{table}

\subsection{Quantitative Results on SYMSOL}

In this section, we present the quantitative results evaluated on SYMSOL, and compare our diffusion-based methods with non-parametric ones. We assess the performance of our score model on $SO(3)$ across various shapes using both ResNet34 and ResNet50 as the backbones, with the results reported in Table~\ref{tab:symsol_result}. Our model demonstrates promising performance, consistently surpassing the contemporary non-parametric baseline models. It is observed that our model, even when based on the less complex ResNet34 backbone, is still able to achieve results that exceed those of the other baselines using the more complex ResNet50 backbone. The average angular errors are consistently below $1$ degree across all shape categories. The performance further improves when employing ResNet50, which emphasizes the potential robustness and scalability of using diffusion models for addressing the pose ambiguity problem. However, it is important to observe that our model with ResNet50 exhibits a slightly reduced performance for the cone shape compared to the ResNet34 variant. This discrepancy can be attributed to our practice of training a single model across all shapes, a strategy that parallels those adopted by Implicit-PDF~\citep{implicitpdf} and HyperPosePDF~\citep{hyperpose}. Such an approach may lead to mutual influences among shapes with diverse pose distributions, and potentially compromise optimal performance for certain shapes. This observation highlights opportunities for future improvements to our model, specifically in enhancing its ability to effectively learn from data spanning various domains. Such endeavors would potentially shed light on the diverse complexities associated with distinct shapes and characteristics.

\begin{figure}
    \centering
    \includegraphics[width=\linewidth]{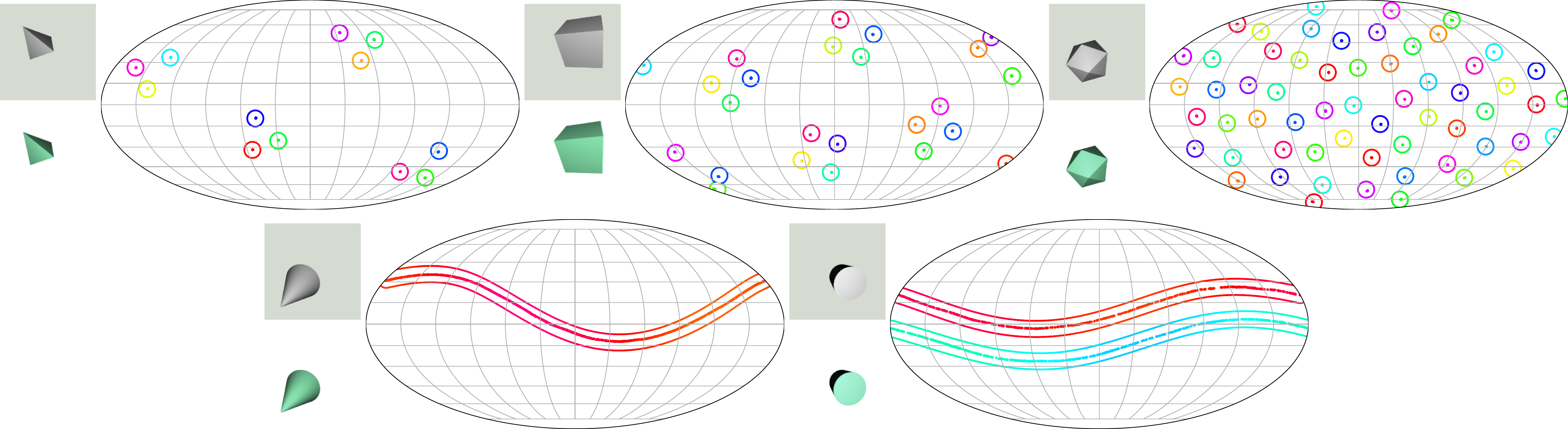}
    \caption{Visualization of our $SE(3)$ diffusion results on SYMSOL-T. Each plot contains $1,000$ sampled poses generated by our model. The first row depicts the densities of discrete symmetrical shapes: (a) tetrahedron, (b) cube, (c) icosahedron, each possessing 12, 24 and 60 discrete symmetries, respectively. The second row presents the densities of continuous symmetrical objects: (d) cone and (e) cylinder, with each shape exhibiting 1 and 2 continuous symmetries, respectively.}
    \label{fig:viz-symsolt}
    \vspace{-1em}
\end{figure}

\begin{figure*}[t]
  \centering
  \includegraphics[width=.9\linewidth]{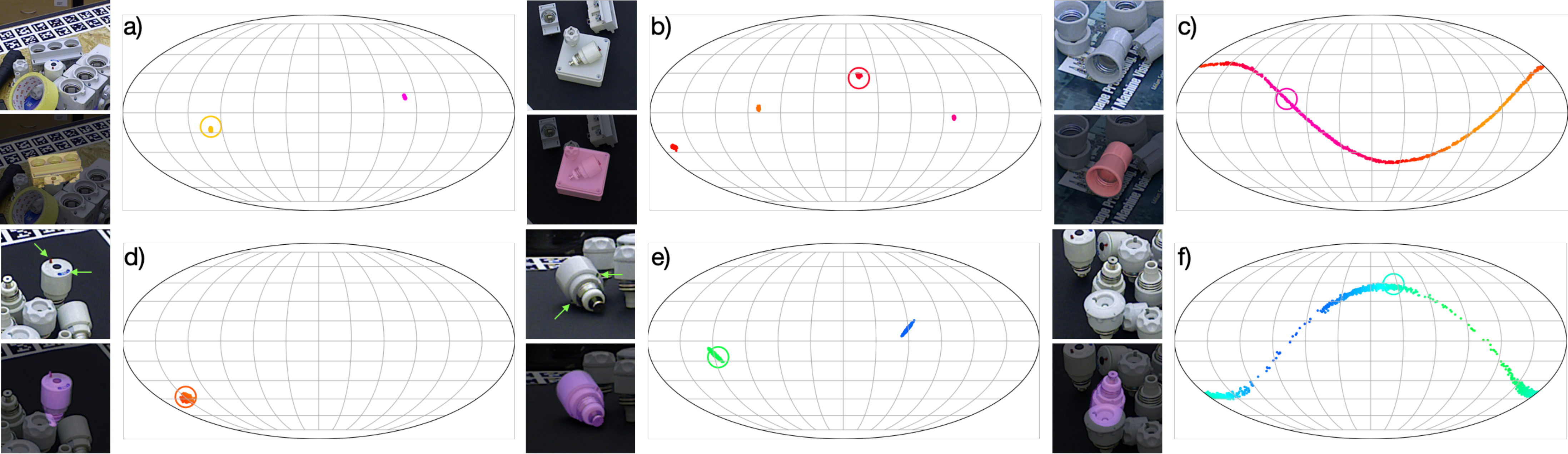}
  \caption{Visualization of our $SE(3)$ diffusion results on T-LESS. In the first row, we present our estimation results of three objects in cluttered scenes: (a) Object 9, characterized by 2 discrete symmetries; (b) Object 27, featuring 4 discrete symmetries; and (c) object 14, possessing 1 continuous symmetries. The second row illustrates pose ambiguities arising from occlusion and self-occlusion, particularly related to Object 4. Notably, this object is annotated with 1 continuous symmetry by human annotator, which does not accurately capture the true ambiguities in certain cases. We explore the scenarios where (d) the object has no symmetry if the top feature is visible; (e) 2 discrete symmetries when the feature is self-occluded, but revealing the two screw holes at the bottom; and (f) 1 continuous symmetry if the screw holes are also occluded by the scene. Each plot contains $1,000$ pose samples from our model. The samples are concentrated on each mode of the distribution, indicating that our models can generate precise rotation estimations across different objects.}
  \label{fig:tless-viz}
  \vspace{-1em}
\end{figure*}

\subsection{Quantitative Results on SYMSOL-T}

We report the quantitative results obtained from the SYMSOL-T dataset evaluation, as shown in Table~\ref{tab:symsolt_result}. The results reveal that our $SE(3)$ and
$R^3SO(3)$ score models outperform the pose regression and iterative regression baselines in terms of estimation accuracy. However, the $R^3SO(3)$ score model encounters difficulty when learning the distribution of the icosahedron shape. In contrast, our $SE(3)$ score model excels in estimating rotation across all shapes and achieves competitive results in translation compared to the $R^3SO(3)$ score model, thus demonstrating its ability to model the joint distribution of rotation and translation. Please note that the $SE(3)$ and $R^3SO(3)$ score models do not rely on symmetry annotations, which distinguish them from the pose regression and iterative regression baselines that leverage symmetry supervision. This supports our initial hypothesis that score models are capable of addressing the pose ambiguity problem in the image domain.
In the comparison between the $R^3SO(3)$ score model and iterative regression, both models employ iterative refinement. However, our $R^3SO(3)$ score model consistently outperforms iterative regression on tetrahedron, cube, cone, and cylinder shapes. The key difference is that iterative regression focuses on minimizing pose errors without explicitly learning the underlying true distributions. In contrast, our $R^3SO(3)$ score model captures different scales of noise, enabling it to learn the true distribution of pose uncertainty and achieve more accurate results.
Regarding translation performance, the $R^3SO(3)$ score model takes the lead over the $SE(3)$ score model. The former's performance can be credited to its assumption of independence between rotation and translation, which effectively eliminates mutual interference. On the other hand, the $SE(3)$ score model learns the joint distribution of rotation and translation, which leads to more robust rotation estimations. The observations therefore support our hypothesis that $SE(3)$ can provide a more comprehensive pose estimation than $R^3SO(3)$. Fig.~\ref{fig:viz-symsolt} show the visualization derived by our model on the $SE(3)$ group.

\begin{table}[t]
    \centering
    \footnotesize
    \caption{Inference time (second per sample) across different denoising steps on the T-LESS dataset.}
    \vspace{-1em}
    \resizebox{\linewidth}{!}{%
    \begin{tabular}{l|c|c|c||c|c|c}
    \specialrule{0.8pt}{0.0ex}{0.2ex}
    \textbf{Methods} & \textbf{Steps} & \textbf{Inference time} & \textbf{FPS} & \textbf{MSPD} & \textbf{MSSD} & \textbf{VSD} \\ 
    \hline
    \multirow{4}{*}{Ours ($R^{3}SO(3)$)}  & 100 & 0.041 & 24 & 85.73 & 52.03 & 48.41  \\ 
             & 50  & 0.021 & 47 & 85.46 & 52.18 & 48.41  \\ 
             & 10  & 0.005 & 188 & 85.57 & 52.25 & 48.77  \\ 
             & 5   & 0.003 & 307 & 85.67 & 53.11 & 49.59 \\ 
    \hline
    \multirow{4}{*}{Ours ($SE(3)$)}    & 100 & 0.050 & 20 & 93.16 & 60.17 & 56.88 \\ 
             & 50  & 0.026 & 38 & 93.00 & 59.96 & 56.64 \\ 
             & 10  & 0.006 & 161 & 92.79 & 60.35 & 57.08 \\ 
             & 5   & 0.004 & 250 & 92.40 & 59.30 & 56.15 \\ 
    \specialrule{0.8pt}{0.2ex}{0.0ex}
    \end{tabular}}
    \label{tab:inference_time}
    \vspace{-1.5em}
\end{table}

\subsection{Quantitative Results on T-LESS}

We evaluate our $SE(3)$ diffusion model on T-LESS, and demonstrate the effectiveness of our approaches in real-world cluttered scenarios. 
In this experiment, a single model with a ResNet34 backbone is trained across 30 T-LESS objects.
We crop the Region of Interest (RoI) confined within bounding boxes from RGB images and employ segmentation masks to isolate the visible parts of objects. To introduce randomness during training while preserving the RoI aspect ratios, we leverage the Dynamic Zoom-In~\citep{cdpn} method. In addition, we apply hard image augmentations~\cite{gdrnet} to the RoIs, including random colors, Gaussian blur, and noise. It is crucial to note that our method assumes the availability of ground truth bounding boxes and segmentation masks for the visible parts of objects. Table~\ref{tab:tless} presents the quantitative results. For comparison, we include GDRNPP~\citep{gdrnet}, a regression-based method that stands as the state-of-the-art approach from the BOP challenge in 2022~\citep{bop2022}. The results indicate that our $SE(3)$ diffusion model outperforms its $R^3SO(3)$ counterpart across all metrics. Furthermore, our $SE(3)$ diffusion model demonstrates superior rotation estimation compared to GDRNPP, albeit with a slightly inferior performance in translation. This discrepancy is attributed to GDRNPP's use of geometry guidance derived from 3D models to enhance depth estimation.  Fig.~\ref{fig:tless-viz} presents the visualization results. Please note that more details are presented in the supplementary material.

\subsection{Inference Time Analysis}
To assess the inference time performance of our models, they are evaluated using the T-LESS dataset and employing JAX~\cite{jax2018github} as the deep learning package. Our experiments are conducted on an AMD Ryzen Threadripper 2990WX CPU and an RTX 2080 Ti GPU. The models, based on the ResNet34 backbone and an input size of 224 x 224 pixels, demonstrate noticeable efficiency across various denoising steps when parametrized on the $SE(3)$ and $R^3SO(3)$ spaces, as detailed in Table~\ref{tab:inference_time}. For $SE(3)$, we achieve up to 250 FPS at minimal denoising steps, while for $R^3SO(3)$, the performance reaches 307 FPS. These results suggest the practical applicability of our models in real-time scenarios.

\vspace{-0.5em}
\section{Conclusion}
\label{sec:conclusion}

In this paper, we presented a novel approach that applies diffusion models to the $SE(3)$ group for object pose estimation, effectively addressing the pose ambiguity issue. Inspired by the correlation between rotation and translation distributions caused by image projection effects, we jointly estimated their distributions on $SE(3)$ for improved accuracy. This is the first work to apply diffusion models to $SE(3)$ in the image domain. To validate it, we developed the SYMSOL-T dataset, which enriches the original SYMSOL dataset with randomly sampled translations. Our experiments confirmed the applicability of our $SE(3)$ diffusion model in the image domain and the advantage of $SE(3)$ parametrization over $R^{3}SO(3)$. Moreover, our experiments on T-LESS exhibits the efficacy of our $SE(3)$ diffusion model in real-world applications.
\section{Acknowledgement}

The authors gratefully acknowledge the support from the National Science and Technology Council (NSTC) in Taiwan under grant numbers MOST 111-2223-E-007-004-MY3, Taiwan. The authors would also like to express their appreciation for the donation of the GPUs from NVIDIA Corporation and NVIDIA AI Technology Center (NVAITC) used in this work. Furthermore, the authors extend their gratitude to the National Center for High-Performance Computing (NCHC) for providing the necessary computational and storage resources.

{
    \small
    \bibliographystyle{ieeenat_fullname}
    \bibliography{main}
}


\clearpage
\setcounter{page}{1}
\maketitlesupplementary

\section{Ablation Studies}
\begin{figure}
	\centering
    \includegraphics[width=\linewidth]{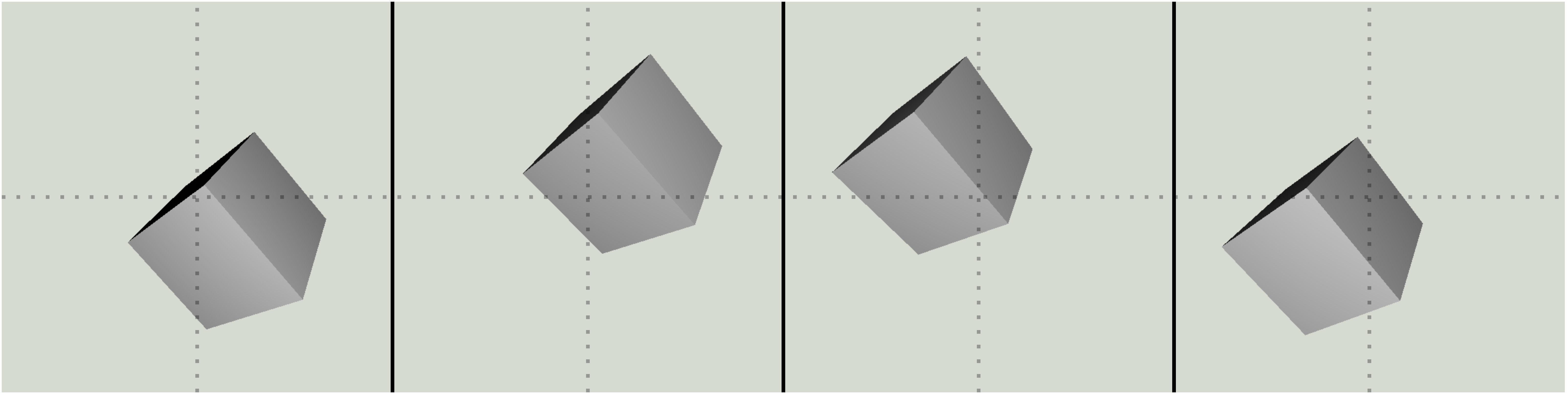}
    \caption{Visualizing pose ambiguity caused by image perspective. The rotations between the four cubes differ by an angle of 15 degrees.}
    \label{fig:ambiguity}
\end{figure}


\begin{figure*}[t]
    \centering
    \includegraphics[width=\textwidth]{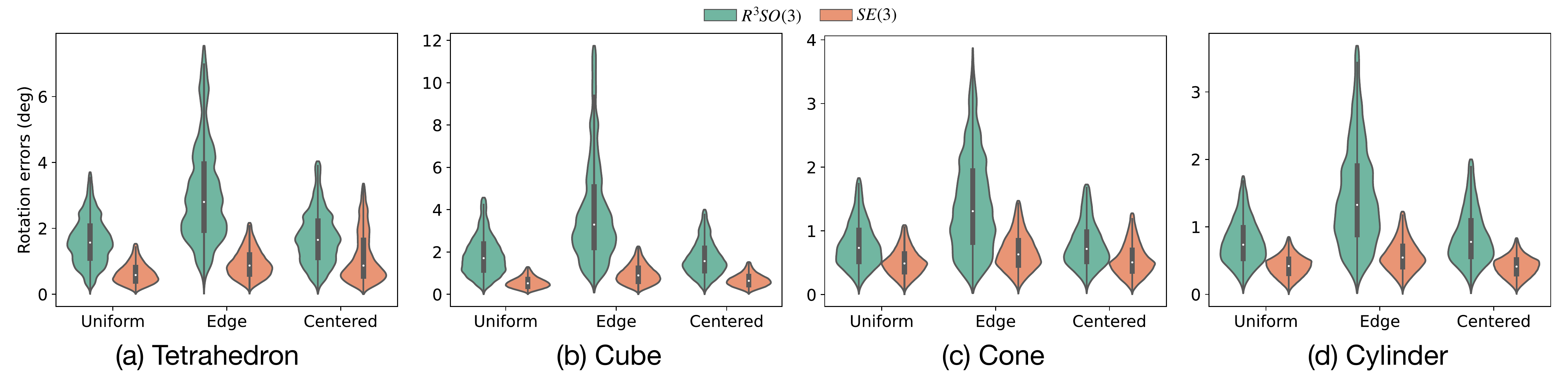}
    \caption{The distribution of angular errors of the $SE(3)$ and $R^3SO(3)$ score models with three configurations and four shapes, in which the width represents the density of data points at a particular range. Please note that the results of $R^3SO(3)$ on \textit{icosa.} are not reported as this model fails to adequately handle this particular shape.}
    \label{fig:perspective-uncertainty}
\end{figure*}

\subsection{Analysis of $SE(3)$ and $R^3SO(3)$ in the Presence of Image Perspective Ambiguity}
\label{sec:image-perspective-ambiguity-exp}
In the realm of pose estimation, the effect of image perspective present a notable challenge. It intertwines rotation and translation in the image space, leading to the phenomenon of pose ambiguity.
Fig.~\ref{fig:ambiguity} exemplifies this through four cubes, each of which appears similarly oriented but actually differs in rotation degrees, complicating model predictions for accurate rotation angles. The parametrizations of $R^{3}SO(3)$ and $SE(3)$ offer different approaches to dealing with this problem. Specifically, $R^{3}SO(3)$ does not factor in the relationship between rotation and translation, whereas $SE(3)$ actively incorporates it into its structure. As a result, it is reasonable to hypothesize that $SE(3)$ might be more capable of mitigating performance degradation stemming from the image perspective effect. This potential advantage of $SE(3)$, further elaborated in Section~\ref{sec:background:transformation}.

To delve deeper into the effects of image perspective on our pose estimation methods, we additionally synthesized three variants of the SYMSOL-T dataset: \textit{Uniform}, \textit{Edge}, and \textit{Centered}. The \textit{Uniform} variant consists of uniformly sampled translations, the \textit{Edge} variant includes translations at the maximum distance from the center, and the \textit{Centered} variant comprises zero translations. Fig.~\ref{fig:perspective-uncertainty} showcases a comparison of the evaluation results for these three variants.  We present the distributions of angular errors made by the $SE(3)$ and $R^3SO(3)$ diffusion models on these dataset variants and four shapes: tetrahedron, cube, cone, and cylinder.   These distributions of angular errors depict the uncertainty of the pose estimations. In line with our hypothesis, the \textit{Edge} variant, which is most influenced by image perspective, exhibits greater uncertainty compared to the \textit{Centered} variant. The \textit{Uniform} variant situates itself between these two. It is evident that both the $R^{3}SO3$ and $SE(3)$ score models demonstrate higher uncertainty on the \textit{Edge} dataset across all shapes, with reduced uncertainty on the \textit{Centered} dataset. The $SE(3)$ score model demonstrates an impressive ability to counter the pose ambiguity introduced by image perspective, a capability that becomes evident when compared with the $R^{3}SO(3)$ score model. The observation therefore confirms our hypothesis that $SE(3)$ does exhibit greater robustness to the ambiguity caused by the image perspective issue.

\begin{table}[t]
\centering
\caption{Evaluation results for various denoising steps applied to score models on $SE(3)$, trained using automatic differentiation and surrogate scores.
}
\label{tab:autograd}
\resizebox{0.48\textwidth}{!}{
\begin{tabular}{l|c|cccccccccc}
\specialrule{0.8pt}{0.0ex}{0.2ex}
\multirow{3}{*}{\textbf{Methods}} & \multirow{3}{*}{\textbf{Steps}} & \multicolumn{10}{c}{\textbf{SYMSOL-T (Spread in degrees $\downarrow$)}} \\ \cline{3-12} 
& & \multicolumn{2}{c|}{\textbf{tet.}} & \multicolumn{2}{c|}{\textbf{cube}} & \multicolumn{2}{c|}{\textbf{icosa.}} & \multicolumn{2}{c|}{\textbf{cone}} & \multicolumn{2}{c}{\textbf{cyl.}} \\ \cline{3-12} 
& & $R$ & \multicolumn{1}{c|}{$t$} & $R$ & \multicolumn{1}{c|}{$t$} & $R$ & \multicolumn{1}{c|}{$t$} & $R$ & \multicolumn{1}{c|}{$t$} & $R$ & $t$ \\
\hline
\multirow{4}{*}{$SE(3)$-autograd} & 100 & 0.60  & \multicolumn{1}{c|}{0.019} & 0.59  & \multicolumn{1}{c|}{0.012} & 0.67 & \multicolumn{1}{c|}{0.012} & 0.58  & \multicolumn{1}{c|}{0.018} & 0.41 & 0.012 \\
& 50 & 0.61 & \multicolumn{1}{c|}{0.019} & 0.61  & \multicolumn{1}{c|}{0.013} & 0.66  & \multicolumn{1}{c|}{0.013} & 0.58  & \multicolumn{1}{c|}{0.019} & 0.41 & 0.013 \\
& 10 & 2.89  & \multicolumn{1}{c|}{0.102} & 3.21  & \multicolumn{1}{c|}{0.113} & 3.24  & \multicolumn{1}{c|}{0.113} & 3.12  & \multicolumn{1}{c|}{0.104} & 3.16 & 0.108 \\
& 5 & 12.93 & \multicolumn{1}{c|}{0.418} & 13.07 & \multicolumn{1}{c|}{0.407} & 10.33 & \multicolumn{1}{c|}{0.302} & 10.83 & \multicolumn{1}{c|}{0.377} & 10.09 & 0.345 \\ 
\hline
\multirow{4}{*}{\shortstack{$SE(3)$-surrogate\\(Ours)}} & 100 & 0.59 & \multicolumn{1}{c|}{0.016} & 0.58  & \multicolumn{1}{c|}{0.011} & 0.64 & \multicolumn{1}{c|}{0.012} & 0.55  & \multicolumn{1}{c|}{0.016} & 0.41 & 0.011 \\
& 50 & 0.56  & \multicolumn{1}{c|}{0.017} & 0.58  & \multicolumn{1}{c|}{0.011} & 0.65 & \multicolumn{1}{c|}{0.012} & 0.54  & \multicolumn{1}{c|}{0.017} & 0.41 & 0.011 \\
& 10 & 0.63  & \multicolumn{1}{c|}{0.017} & 0.70  & \multicolumn{1}{c|}{0.012} & 1.71  & \multicolumn{1}{c|}{0.015} & 0.56  & \multicolumn{1}{c|}{0.019} & 0.43 & 0.014 \\
& 5 & 1.22  & \multicolumn{1}{c|}{0.024} & 2.00  & \multicolumn{1}{c|}{0.028} & 5.31  & \multicolumn{1}{c|}{0.048} & 0.72  & \multicolumn{1}{c|}{0.035} & 0.62 & 0.031  \\
\specialrule{0.8pt}{0.2ex}{0.0ex}
\end{tabular}}
\end{table}

\subsection{Performance Analysis: Surrogate Score versus Automatically Differentiated True Score}
\label{sec:surrogate_vs_autograd}
To evaluate our hypothesis concerning convergence speed, we compare two versions of our score model. The first version, termed $SE(3)$-surrogate, is trained with the \textit{surrogate score} described in Eq.~(\ref{eq:surrogate_score}). The second version, termed as $SE(3)$-autograd, is trained with the \textit{true score} described in Eq.~(\ref{eq:score_jacob}) and calculated by automatic differentiation as described in Section~\ref{sec:autodiff}.  We trained both estimators and evaluated their performance using different steps of denoising process. The results are reported in Table~\ref{tab:autograd}. Our findings show that when a larger number of denoising steps (e.g., $100$ steps) are used, both score models produce comparable results. However, the performance of $SE(3)$-autograd significantly declines in comparison to $SE(3)$-surrogate when the number of sampling steps decreases from $50$ to $10$ and then to $5$.  This performance drop is due to the curved manifold represented by the $SE(3)$ parametrization, which can result in the score vector not consistently pointing towards the noise-free data. These results substantiate our hypothesis, and suggest that the application of the \textit{surrogate score} can lead to faster convergence than the use of the \textit{true score} calculated through automatic differentiation.

\begin{table}[t]
    \caption{Comparison with other diffusion-based approaches.}
    \label{tab:diffusion_comparison}
    \centering
    \resizebox{0.47\textwidth}{!}{
    \begin{tabular}{c|c|c|c|c|c|c|c|c}
        \specialrule{0.8pt}{0.0ex}{0.2ex}
        \multirow{2}{*}{\textbf{Methods}} & \multirow{2}{*}{\textbf{Distribution}} & \multirow{2}{*}{\textbf{Loss}} & \multicolumn{6}{c}{\textbf{SYMSOL (Spread in degrees $\downarrow$)}} \\ \cline{4-9}
        \STstrut & & & \textbf{Avg.} & \textbf{tet.} & \textbf{cube} & \textbf{icosa.} & \textbf{cone} & \textbf{cyl.} \\ \hline
        \STstrut Leach \etal~\citep{leach2022} & $\mathcal{IG}_{SO(3)}$ & DDPM & 0.63 & 0.59 & 0.65 & 0.75 & 0.73 & 0.41 \\ 
        \STstrut Jagvaral \etal~\citep{jagvaral2023} & $\mathcal{IG}_{SO(3)}$ & MLE & 30.45 & 12.21 & 15.18 & 28.76 & 86.77 & 9.35 \\ \hline
        \STstrut Ours w/o fourier & $\mathcal{IG}_{SO(3)}$ & DSM & 1.18 & 0.52 & 0.77 & 3.97 & \textbf{0.32} & \textbf{0.32} \\
        \STstrut Ours w/o fourier & $\mathcal{N}_{SO(3)}$ & DSM & 0.51 & 0.50 & 0.46 & 0.91 & 0.33 & 0.34 \\ 
        \STstrut Ours & $\mathcal{N}_{SO(3)}$ & DSM & \textbf{0.42} & \textbf{0.43} & \textbf{0.44} & \textbf{0.52} & 0.35 & 0.35 \\
        \specialrule{0.8pt}{0.2ex}{0.0ex}
    \end{tabular}}
\end{table}
\subsection{Comparison of Diffusion Models on $SO(3)$}
In this experiment, we further compare our $SO(3)$ score model with the diffusion models proposed by~\citep{leach2022} and~\citep{jagvaral2023} using the SYMSOL dataset. While these studies do not specifically address object pose estimation, we have adapted their methods to fit within our framework.
The authors of~\citep{leach2022} extend the DDPM~\cite{ddpm} to $SO(3)$ using an analogy approach. They employ an $SO(3)$ variant of DDPM loss during the training process.
On the another hand, the authors of~\citep{jagvaral2023} reformulate the SGM~\cite{ncsn} to apply it to the $SO(3)$ space and proposed to train with maximum log-likelihood loss (MLE). The results of these comparisons are presented in Table~\ref{tab:diffusion_comparison}. Our analysis shows that the models employing DDPM or Denoising Score Matching (DSM) losses can learn the distributions on $SO(3)$ effectively, while the model employing MLE loss fails. When comparing our score models with different distributions, we can observe that the one with $\mathcal{N}_{SO(3)}$ performs better than it $\mathcal{IG}_{SO(3)}$ counterpart. Furthermore, when incorporating the Fourier-based conditioning descreibed in Section~\ref{sec:methodology:proposed_framework}, our score model can achieve the best performance on SYMSOL. This suggests that Fourier-based conditioning enhances our models ability to learn pose distributions.

\begin{table}[t]
    \centering
    \caption{Evaluation results on T-LESS (30 objects).}
    \resizebox{\linewidth}{!}{%
    \begin{tabular}{c|c|c|c||c|c|c||c|c|c}
    \specialrule{0.8pt}{0.0ex}{0.2ex}
    \multirow{2}{*}{\textbf{Objects}} & \multicolumn{9}{c}{\textbf{T-LESS (Accuracy \% $\uparrow$)}} \\ \cline{2-10}
    & \textbf{MSPD} & \textbf{MSSD} & \textbf{VSD} & \textbf{R@2} & \textbf{R@5} & \textbf{R@10} & \textbf{T@2} & \textbf{T@5} & \textbf{T@10} \\
    \hline
    \STstrut     1     &       90.05 &           32.29 &          29.60 &   38.22 &   78.20 &    89.10 &   40.78 &   72.14 &    89.50  \\
    \STstrut     2     &       92.22 &           35.56 &          31.73 &   48.07 &   85.49 &    92.97 &   42.63 &   73.92 &    91.61  \\
    \STstrut     3     &       97.55 &           47.29 &          43.88 &   52.86 &   92.45 &    98.70 &   60.42 &   90.10 &    96.88  \\
    \STstrut     4     &       92.27 &           48.84 &          46.07 &   44.28 &   86.36 &    93.43 &   52.86 &   85.52 &    95.12  \\
    \STstrut     5     &       96.32 &           76.47 &          74.18 &   49.47 &   91.58 &    96.84 &   81.05 &   95.79 &    98.95  \\
    \STstrut     6     &       98.57 &           78.06 &          75.71 &   60.20 &   92.86 &    97.96 &   84.69 &   95.92 &    97.96  \\
    \STstrut     7     &       93.96 &           85.44 &          80.50 &   54.80 &   94.00 &    98.00 &   80.80 &   95.60 &    99.60  \\
    \STstrut     8     &       90.40 &           86.53 &          79.49 &   44.67 &   93.33 &    98.00 &   70.00 &   96.00 &    98.00  \\
    \STstrut     9     &       96.54 &           84.15 &          79.46 &   47.15 &   93.09 &    97.97 &   82.93 &   97.56 &    99.59  \\
    \STstrut    10     &       98.39 &           68.88 &          63.35 &   50.35 &   90.91 &    99.30 &   72.03 &   95.10 &    99.30  \\
    \STstrut    11     &       95.20 &           57.26 &          51.52 &   25.14 &   77.71 &    91.43 &   68.00 &   93.14 &    98.86  \\
    \STstrut    12     &       96.76 &           62.23 &          56.47 &   38.85 &   87.05 &    95.68 &   64.75 &   93.53 &    97.12  \\
    \STstrut    13     &       99.36 &           47.79 &          44.89 &   70.00 &   96.43 &   100.00 &   62.86 &   91.43 &    99.29  \\
    \STstrut    14     &       97.60 &           63.36 &          60.05 &   71.92 &   95.21 &    98.63 &   71.92 &   94.52 &    97.26  \\
    \STstrut    15     &       97.95 &           59.93 &          57.72 &   73.97 &   97.95 &    98.63 &   69.86 &   93.15 &    98.63  \\
    \STstrut    16     &       97.34 &           61.81 &          59.40 &   67.02 &   96.28 &    97.87 &   76.06 &   92.55 &    97.87  \\
    \STstrut    17     &       98.56 &           82.19 &          78.47 &   78.08 &   98.63 &   100.00 &   85.62 &   96.58 &    97.26  \\
    \STstrut    18     &       83.42 &           72.33 &          75.22 &   16.44 &   59.59 &    78.77 &   82.19 &   93.84 &    95.21  \\
    \STstrut    19     &       94.03 &           64.71 &          60.83 &   28.80 &   79.58 &    94.76 &   70.16 &   92.67 &    97.91  \\
    \STstrut    20     &       88.71 &           61.62 &          54.42 &   22.92 &   70.83 &    92.08 &   65.00 &   90.00 &    97.92  \\
    \STstrut    21     &       80.06 &           58.00 &          56.74 &   37.71 &   72.57 &    77.71 &   68.57 &   84.57 &    90.86  \\
    \STstrut    22     &       83.94 &           59.20 &          58.82 &   29.26 &   72.34 &    84.57 &   70.21 &   90.96 &    96.28  \\
    \STstrut    23     &       92.58 &           78.06 &          73.75 &   25.00 &   78.63 &    94.76 &   72.98 &   96.77 &    98.39  \\
    \STstrut    24     &       96.98 &           62.29 &          59.27 &   56.77 &   95.31 &    97.40 &   65.10 &   92.71 &    98.96  \\
    \STstrut    25     &       94.84 &           74.84 &          71.48 &   48.42 &   91.58 &    97.89 &   78.95 &   95.79 &    97.89  \\
    \STstrut    26     &       97.17 &           81.41 &          78.73 &   49.49 &   97.98 &    98.99 &   90.91 &   96.97 &    98.99  \\
    \STstrut    27     &       89.69 &           79.90 &          75.27 &   33.33 &   81.25 &    94.79 &   82.29 &   93.75 &    97.92  \\
    \STstrut    28     &       88.12 &           73.12 &          72.58 &   39.58 &   78.65 &    90.62 &   75.52 &   91.15 &    95.31  \\
    \STstrut    29     &       95.82 &           84.90 &          83.78 &   53.06 &   90.82 &    97.96 &   84.69 &   96.94 &    98.98  \\
    \STstrut    30     &       97.85 &           69.86 &          67.50 &   60.42 &   91.67 &    98.61 &   77.78 &   92.36 &    97.22  \\
    \hline
    \STstrut  Avg(30)  &       93.16 &           60.17 &          56.88 &   47.21 &   86.94 &    94.78 &   71.72 &   92.03 &    97.15  \\
    \specialrule{0.8pt}{0.2ex}{0.0ex}
    \end{tabular}}
    \label{tab:tless-full}
\end{table}

\subsection{Full Evaluation Results on T-LESS}
Table~\ref{tab:tless-full} presents the evaluation results of our $SE(3)$ diffusion model on each T-LESS object. Please note that a single model with ResNet34 backbone is trained across thirty T-LESS objects. More visualization results are presented in Fig.~\ref{fig:tless-viz-sup}.

\begin{table}[t]
    \centering
    \caption{Evaluation results on T-LESS (Average of 30 objects).}
    \resizebox{\linewidth}{!}{%
    \begin{tabular}{l|c|c|c||c|c|c||c|c|c}
    \specialrule{0.8pt}{0.0ex}{0.2ex}
    \multirow{2}{*}{\textbf{Methods}} & \multicolumn{9}{c}{\textbf{T-LESS (Accuracy \% $\uparrow$)}} \\ \cline{2-10}
    & \textbf{MSPD} & \textbf{MSSD} & \textbf{VSD} & \textbf{R@2} & \textbf{R@5} & \textbf{R@10} & \textbf{T@2} & \textbf{T@5} & \textbf{T@10} \\
    \hline
    \STstrut GDRNPP~\citep{gdrnet} & 90.17 & 75.06 & 67.60 & 21.60 & 71.18 & 90.56 & 90.31 & 96.09 & 98.10 \\
    \hline
    \STstrut Ours ($R^{3}SO(3)$) & 85.73 & 52.03 & 48.41 & 27.98 & 72.42 & 89.26 & 60.37 & 79.75 & 89.62  \\ 
    \STstrut Ours ($SE(3)$) & \textbf{93.16} & 60.17 & 56.88 & \textbf{47.21} & \textbf{86.94} & \textbf{94.78} & 71.72 & 92.03 & 97.15 \\
    \specialrule{0.8pt}{0.2ex}{0.0ex}
    & \textbf{x@2} & \textbf{x@5} & \textbf{x@10} & \textbf{y@2} & \textbf{y@5} & \textbf{y@10} & \textbf{z@2} & \textbf{z@5} & \textbf{z@10} \\
    \hline
    \STstrut GDRNPP~\citep{gdrnet} & 98.12 & 98.84 & 99.47 & 98.56 & 99.35 & 99.59 & 91.21 & 96.67 & 98.56 \\
    \hline
    \STstrut Ours ($R^{3}SO(3)$) & 98.00 & 99.66 & 99.92 & 96.46 & 99.82 & 99.99 & 61.68 & 80.23 & 89.94 \\ 
    \STstrut Ours ($SE(3)$) & 99.20 & 99.63 & 99.88 & 99.19 & 99.81 & 99.99 & 73.33 & 92.51 & 97.33 \\
    \specialrule{0.8pt}{0.2ex}{0.0ex}
    \end{tabular}}
    \label{tab:tless-translation}
\end{table}

\subsection{Translation Analysis on T-LESS}
In this section, we further analyze the error sources of our $SE(3)$ diffusion model and GDRNPP~\citep{gdrnet}. The translation accuracies on $x$, $y$ and $z$ axes are reported in Table~\ref{tab:tless-translation}. It can be observed that the $SE(3)$ diffusion model is able to predict the $x$ and $y$ translations as accurate as GDRNPP. However, the $SE(3)$ diffusion model exhibits a slightly less effective performance in predicting the depth value $z$ compared to GDRNPP. This is because GDRNPP employs geometry guidance~\citep{gdrnet} by the reconstructed 3D models of the objects to enhance depth estimation, while our $SE(3)$ diffusion model exclusively depends on RGB inputs and ground truth poses for supervision. Nevertheless, these results still highlight the significant potential of our diffusion models to compete with contemporary state-of-the-art methods on the real-world datasets.

\subsection{Failure Analysis on T-LESS}
The failure cases are provided in Fig.~\ref{fig:tless-viz-sup}. In Fig.~8~(a), our approach predicts the pose as exhibiting one continuous symmetry. However, in reality, there should be only six discrete symmetries. This presents a failure case arising from the objective of probabilistic modeling, which aims to approximate the distribution across the entire space. Our assumption regarding the possible reasons is twofold: (a) we fit one model to multiple objects, which may have difficulty representing and learning all the distributions accurately, as they may interfere with each other; (b) another limitation of our diffusion-based approach is its reliance on a sufficient volume of data samples. Without these, it could fail to accurately model the correct distribution of poses.

\section{Additional implementation Details}
\label{sec:add-experimental-details}

\subsection{Isotropic Gaussian on $SO(3)$}

Isotropic Gaussian on $SO(3)$~\citep{nikolayev1970normal}, denoted as $\mathcal{IG}_{SO(3)}$, is a heat kernel that can be used to model the distribution on $SO(3)$ rotation space, which has the following form:
\begin{equation}
{\footnotesize
    \begin{split}
    f_\epsilon(\phi)=\lim_{N\to\infty}\sum^N_{\ell=0}(2\ell+1)e^{-\epsilon\ell(\ell+1)}\frac{\sin((2\ell+1)\phi/2)}{\sin(\phi/2)},
    \end{split}
}
\label{eq:igso3_infinite}
\end{equation}
where $\phi \in [0, \pi]$ is the rotation angle and $\epsilon > 0$ is the concentration parameter. Note that a normalizing factor $Z(\phi) = (1-\cos(\phi))/\pi$ is applied to this distribution. For an $\epsilon \ll 1$, this infinite series converge slowly and could lead to inefficient computation. In the previous literature, the authors in~\citep{yim2023} proposed to truncate the series by letting $N=2000$, while the authors in~\citep{jagvaral2023} attempted to use another closed-form approximation, expressed as follows:
\begin{equation}
{\footnotesize
    \begin{split}
    f_\epsilon(\phi)&\approx\sqrt{\pi}\epsilon^{-\frac{3}{2}}e^{\frac{\epsilon}{4}-\frac{(\phi/2)^2}{\epsilon}} \\
    &\cdot\left(\frac{\phi-e^{-\frac{\pi^2}{\epsilon}}\left((\phi-2\pi)e^{\frac{\pi\phi}{\epsilon}}+(\phi+2\pi)e^{-\frac{\pi\phi}{\epsilon}}\right)}{2\sin(\phi/2)}\right).
    \end{split}
}
\label{eq:igso3_approx}
\end{equation}
As shown in~\citep{matthies1988normal}, this approximation closely aligns with Eq.~(\ref{eq:igso3_infinite}) when $\epsilon < 1$. To draw samples from this distribution, a common approach is to utilize the inverse transform sampling. The steps are described as follows. First, a sample is drawn from a uniform distribution within $[0, \pi]$. Subsequently, the cumulative distribution function (CDF) of $\mathcal{IG}_{SO(3)}$ is calculated for inverse sampling. The sampling procedure is described in Listing~\ref{lst:igso3}.

Unfortunately, $\mathcal{IG}_{SO(3)}$ still exists several drawbacks. The main concern is the intractability of the inverse CDF for $\mathcal{IG}_{SO(3)}$, which necessitates interpolation in the calculation of inverse sampling. Moreover, numerical instability could arise during the inverse sampling when $\epsilon$ is close to zero. As a result, this distribution is not suitable for applications that require precise computations. Therefore, the proposed method opt to utilize an alternative distribution to enhance performance and reliability.

\begin{figure}[t]
\centering
\begin{minipage}{\linewidth}
\begin{lstlisting}[basicstyle=\scriptsize,label={lst:igso3},language=Python,caption=Isotropic Gaussian $SO(3)$ in JAX.]
from math import pi
from jaxlie import SO3
import jax
import jax.numpy as jnp

def normalize(v):
    return v / jnp.linalg.norm(v)


def rsub(y:SO3, x:SO3):
    return (x.inverse() @ y).log()

# geodesic distance
def geodesic(y:SO3, x:SO3):
    return jnp.linalg.norm( rsub(y, x) )

# Ep. (15)
def f_igso3(phi, scale):
    eps = scale ** 2
    return 0.5 * jnp.sqrt(jnp.pi) * (eps**-1.5) \
      * jnp.exp((eps-(phi**2/eps))/4) / jnp.sin(phi/2) \
      * (phi-((phi-2*pi)*(jnp.exp(pi*(phi-pi)/eps))) \
        + (phi+2*pi)*(jnp.exp(-pi*(phi+pi)/eps)))) 

def cdf(steps=1024):
    x = jnp.linspace(0.0, 1.0, steps) * pi
    y = (1-jnp.cos(x)) / pi * f_igso3(x)
    y = jnp.cumsum(y) * pi / steps
    return y / y.max(), x

# Inverse transform sampling
def sample(seed):
    y, x = cdf_igso3()
    key1, key2 = jax.random.split(seed, 2)
    rnd = jax.random.uniform((), key=key1)
    ang = jnp.interp(rnd, y, x)
    axis = jnp.random.normal((3,), key=key2)
    axis = normalize(axis)
    tan = ang[..., jnp.newaxis] * axis
    return SO3.exp(tan)

# log-likelihood IG_SO3
def log_prob(x:SO3, mu:SO3, scale):
    phi = geodesic(mu, x)
    prob = f_igso3(phi, scale)
    return jnp.log(prob)

\end{lstlisting}
\end{minipage}
\end{figure}

\subsection{Concentrated Gaussian on $SO(3)$}
Concentrated Gaussian distribution~\citep{chirikjian2014gaussian, barfoot2014} is a distribution that used for modeling the density on Lie groups. We denote such distribution as $\mathcal{N}_{\mathcal{G}}$, where $\mathcal{G}$ implies specifically applying it on Lie group $\mathcal{G}$. This distribution usually assumes that the noises $z\sim \mathcal{N}(\mathbf{0}, \Sigma)$ are relatively small compared to the domain of the distribution and concentrated around zero in the corresponding vector space. By the definition of multivariate Gaussian distribution, the probability density of $z \in \mathbb{R}^\kappa$ is described as follows:
\begin{equation}
{\footnotesize
\begin{split}
    p_\Sigma(z)
        &:= \mathcal{N}(\mathbf{0}, \Sigma) \triangleq\frac{1}{\sqrt{(2\pi)^\kappa|\Sigma|}}\exp\left(-\frac{1}{2}z^\top\Sigma^{-1} z\right),
\end{split}
}
\label{eq:multi-gaussian}
\end{equation}
where $\Sigma\in\mathbb{R}^{\kappa\times\kappa}$ is the covariance matrix. Assuming that $X, Y \in \mathcal{G}$ and $z \in \mathfrak{g}$, and given the relation $Y = X\Exp(z)$, the inverse relation can be expressed as $z = \Log(X^{-1}Y)$. Substituting this into Eq.~(\ref{eq:multi-gaussian}) results in a concentrated Gaussian on $\mathcal{G}$ centered at $X$. This result corresponds to Eq.~(\ref{eq:perturbation_lie}) in our main paper and can be expressed as follows:
\begin{equation}
{\scriptsize
\begin{split}
p_\Sigma(Y|X) &:= \mathcal{N}_{\mathcal{G}}(Y;X,\Sigma) \\
              &\triangleq \frac{1}{\zeta(\Sigma)}\exp\left(-\frac{1}{2}\Log(X^{-1}Y)^\top\Sigma^{-1}\Log(X^{-1}Y)\right),
\end{split}
}
\end{equation}
where $\zeta(\Sigma)$ is the normalizing factor. To draw samples from this distribution, it is accomplished by first drawing a random variable from the normal distribution $z \sim \mathcal{N}(\mathbf{0}, \Sigma)$. Subsequently, $z$ is applied to the center parameter $X$ to yield $Y = X\Exp(z)$. The sampling procedure is detailed in Listing~\ref{lst:gauss}. The primary advantage of this distribution is its elimination of the need for approximation and inverse sampling. Due to its simplicity, this method has been extensively utilized in prior literature for modeling the distribution on $SO(3)$~\citep{chirikjian2014gaussian}, $SE(3)$~\citep{barfoot2014,urain2022} and manifolds~\citep{riemannian}. 

\begin{figure}[t]
\centering
\begin{minipage}{\linewidth}
\begin{lstlisting}[basicstyle=\scriptsize,label={lst:gauss},language=Python,caption=Concentrated Gaussian $SO(3)$ in JAX.]
from math import pi
from jaxlie import SO3
import jax
import jax.numpy as jnp

def sample(seed, scale):
    tan = jax.random.normal(shape=n+(3,), key=seed)
    tan = tan * scale
    return SO3.exp(tan)

# log-likelihood concentrated Gaussian
def log_prob(x:SO3, mu:SO3, scale):
    var = (scale ** 2)
    log_sc = jnp.log(scale)
    nm = jnp.log(jnp.sqrt(2 * pi))
    z = rsub(mu, x)
    return -((z ** 2) / (2 * var) - log_sc - nm).sum()

\end{lstlisting}
\end{minipage}
\end{figure}

\begin{figure}[t]
\centering
\begin{minipage}{\linewidth}
\begin{lstlisting}[basicstyle=\scriptsize,label={lst:true_score},language=Python,caption=Calculation of Stein scores using automatic differentiation.]
from jaxlie import SO3, SE3
import jax
import jax.numpy as jnp

Lie = SO3  # Specify Lie groups
    
# Eq. (25)
def calc_score(y, x, sigma=1.0):
    return jax.grad(
        lambda tau: log_prob(
            Lie.exp(y) @ Lie.exp(tau),
            Lie.exp(x),
            sigma
        )
    )(jnp.zeros(Lie.tangent_dim))
    # tangent_dim=3 for SO3, 6 for SE3

\end{lstlisting}
\end{minipage}
\end{figure}
\label{sec:autodiff}
\subsection{Calculation of Stein Scores Using Automatic Differentiation in JAX}

As stated by~\cite{jagvaral2023}, the Stein scores can be computed as follows:
\begin{equation}
    \nabla_Y \log p_\Sigma(Y|X) = \left.\frac{\partial }{\partial k} \log p_\Sigma(Y\text{Exp}(k\tau)|X)\right|_{k=0},
    \label{eq:score_autograd}
\end{equation}
where $k\in\mathbb{R}$, $\tau\in \mathfrak{g}$, and $k\tau$ indicates a small perturbation on $\mathcal{G}$. In practice, this can be computed by automatic differentiation. Listing~\ref{lst:true_score} demonstrates our implementation based on \textit{JAX}~\cite{jax2018github} and \textit{jaxlie}~\cite{yi2021iros}.

\SetKwInput{Require}{Require}

\begin{table}[t]
\scalebox{1.0}{
    \begin{minipage}[t]{1.0\linewidth}
    \centering
    \removelatexerror
    \begin{algorithm}[H]
    \caption{Training a Score Model using Denoising Score Matching on $\mathcal{G}$}
    %
    \Require{$s_{\boldsymbol{\theta}}$, $\{\sigma_i\}^L_{i=0}$, $p_\text{data}$}
    \BlankLine
    \For{$j\in\{0,\dots,N_\text{iter}-1\}$}{
        $i\sim \mathcal{U}(0, L-1)$
        
        $X\sim p_\text{data}(X)$
        
        $\tilde{X}=X\Exp(z),~z\sim \mathcal{N}(0,\sigma^2_iI)$
        
        $\ell_\theta=\| s_\mathbf{\theta}(\tilde{X},\sigma_i) - \tilde{s}_X(\tilde{X},\sigma_i) \|^2_2$
        
        $\theta \gets \text{optimize}(\theta, \ell_\theta)$
    }
    
    \label{alg:training}
    \end{algorithm}
    \end{minipage}}
\end{table}

\begin{table}[t]
    \scalebox{1.0}{
    \begin{minipage}[t]{1.0\linewidth}
    \centering
    \removelatexerror
    \begin{algorithm}[H]
    \caption{Sampling Through Geodesic Random Walk on $\mathcal{G}$}
    \Require{$s_{\boldsymbol{\theta}}$, $\{\sigma_i\}^L_{i=0}$, $\{\epsilon_i\}^L_{i=0}$, $\tilde{X}_0$}
    \BlankLine
    \For{$i \in \{0, \dots, L-1\}$}{
        $z_i\sim \mathcal{N}(0, I)$ 
        
        $\tilde{X}_{i+1} = \tilde{X}_{i}\text{Exp}(\epsilon_i s_\theta(\tilde{X}_i, \sigma_i) + \sqrt{2\epsilon_i}z^m_i)$
    }
    \Return{$\tilde{X}_{L}$}
    \label{alg:sampling}
    \end{algorithm}
    \end{minipage}}
\end{table}

\subsection{Algorithms}

The algorithms used for our training and sampling procedures are presented in Algorithms~\ref{alg:training} and~\ref{alg:sampling}, respectively. The notations employed conform to those detailed in the main manuscript.

\subsection{Datasets}

The SYMSOL-T dataset contains 250k images of five symmetric, texture-less three-dimensional objects. Following the structure of SYMSOL~\cite{implicitpdf}, each shape has 45k training images and 5k testing images. The dataset ensures that translations over the $x$, $y$, and $z$ axes are uniformly sampled within the range of $[-1, 1]$. In the experiments examining image perspective ambiguity in Section~\ref{sec:image-perspective-ambiguity-exp}, each of the dataset variants (i.e., \textit{Uniform}, \textit{Edge}, and \textit{Centered}) comprises 200 images per shape. Our analysis is performed based on 1k randomly generated poses from our score models for each image.

\subsection{Hyperparameters}

In our experiments, we utilize a pre-trained ResNet34 model~\citep{resnet} as the standard backbone across all methods, unless explicitly stated otherwise. During training, we sample a batch containing 16 images and the corresponding ground truth poses in each iteration. Each of these samples is perturbed to generate 256 random poses, resulting in 4,096 noisy samples. The proposed score-based model is then trained for 400k steps to denoise these samples. In the SYMSOL-T experiments, the pose regression approach is trained for 400k steps. Meanwhile, the iterative regression and both our $R^{3}SO(3)$ and $SE(3)$ score models are subjected to an extended training duration of 800k steps. In the T-LESS experiments, the size of the batch is increased to 32. The score-based model is trained for 400k steps. We employ the Adam optimizer~\citep{adam} with an initial learning rate set at $10^{-4}$. During the latter half of the training schedule, we apply an exponential decay, which lowers the learning rate to $10^{-5}$. For the diffusion process, we use a linear noise scheduling approach that ranges from $10^{-4}$ to $1.0$, divided into 100 discrete steps.

\begin{table}[h]
\centering
\caption{Hyperparameters.}
\label{tab:hyperparameters}
\resizebox{\linewidth}{!}{
\centering
\begin{tabular}{c|c|c|c}
    \specialrule{0.8pt}{0.0ex}{0.2ex}
    Hyperparameters & SYMSOL & SYMSOL-T & T-LESS\\
    \hline
    Learning rate & $\left[10^{-4}, 10^{-5}\right]$ & $\left[10^{-4}, 10^{-5}\right]$ & $\left[10^{-4}, 10^{-5}\right]$\\
    Batch size & 16 & 16 & 32\\
    Number of noisy samples & 256 & 256 & 256\\
    Training steps & 400k & 800k & 400k\\
    Optimizer & Adam & Adam & Adam\\
    \hline
    Noise scale & $\left[10^{-4}, 1.0\right]$ & $\left[10^{-4}, 1.0\right]$ & $\left[10^{-4}, 1.0\right]$\\
    Denoising steps & 100 & 100 & 100\\
    Number of MLP blocks & 1 & 1 & 1\\
    \specialrule{0.8pt}{0.2ex}{0.0ex}
\end{tabular}
}
\end{table}

\subsection{Evaluation Metrics}

In the SYMSOL experiments, we adopt the minimum angular distance, measured in degrees, between a set of ground truth equivalent rotations and the estimated rotations as the evaluation metric. For the SYMSOL-T experiments, we incorporate the Euclidean distance between the ground truth and the estimated translations as our metric to evaluate the accuracy of translation. Each of these distance metrics is computed per sample, and we report their averages over all samples in our results. In the T-LESS experiments, we adopt three standard metrics used in the BOP challenge~\citep{bop-challenge}: Maximum Symmetry-Aware Projection Distance (MSPD), Maximum Symmetry-Aware Surface Distance (MSSD), and Visible Surface Discrepancy (VSD).

\subsection{Visualization of SYMSOL-T Results}
\label{sec:visualization}

\begin{figure*}[b]
    \centering
    \includegraphics[width=.8\textwidth]{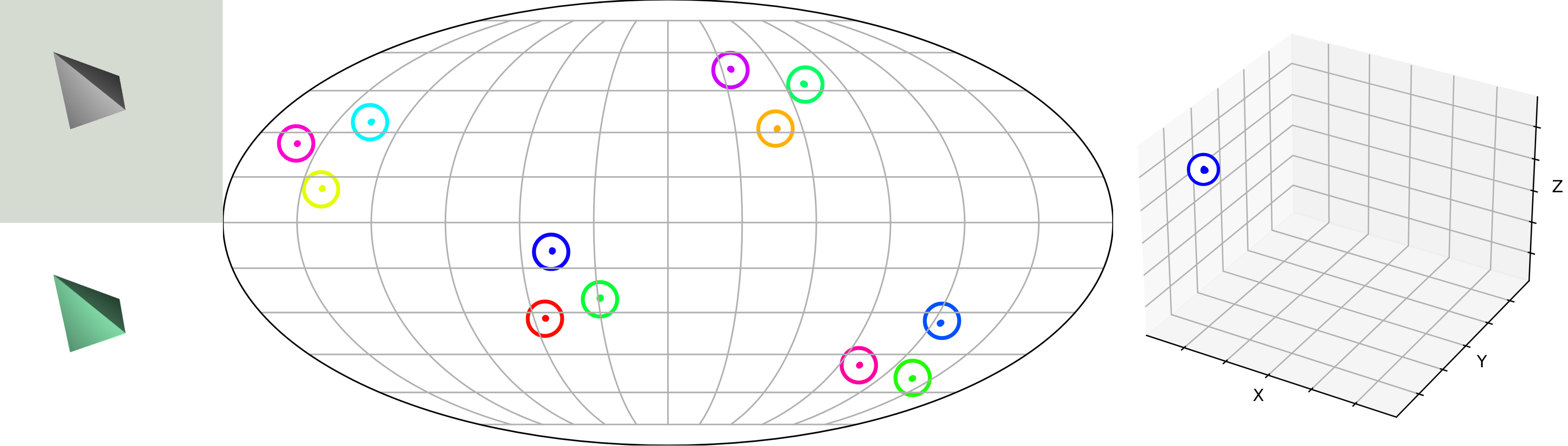}
    \includegraphics[width=.8\textwidth]{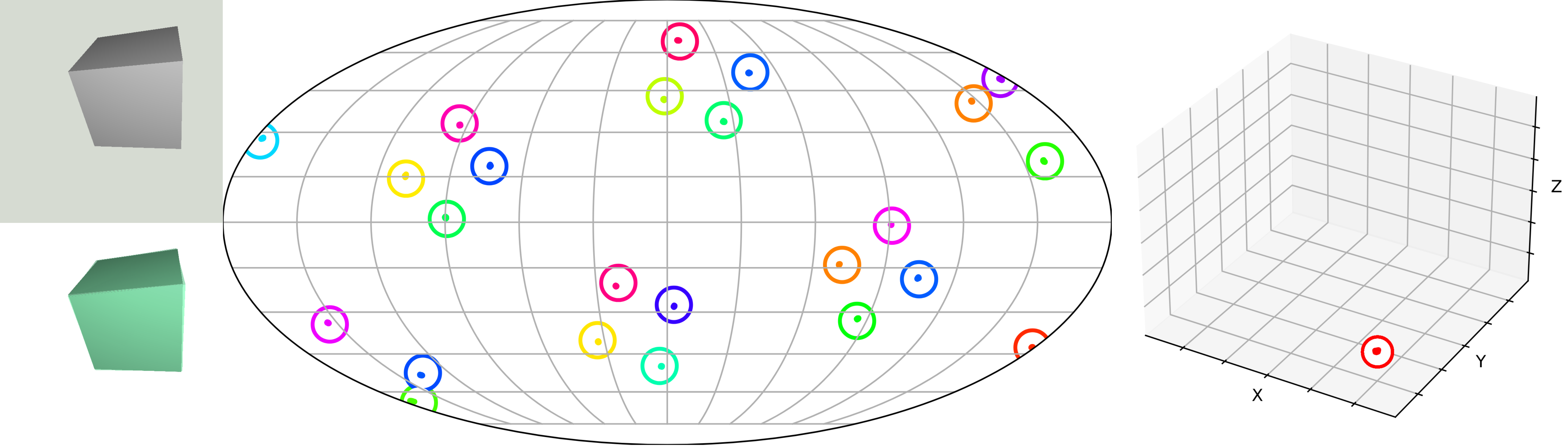}
    \includegraphics[width=.8\textwidth]{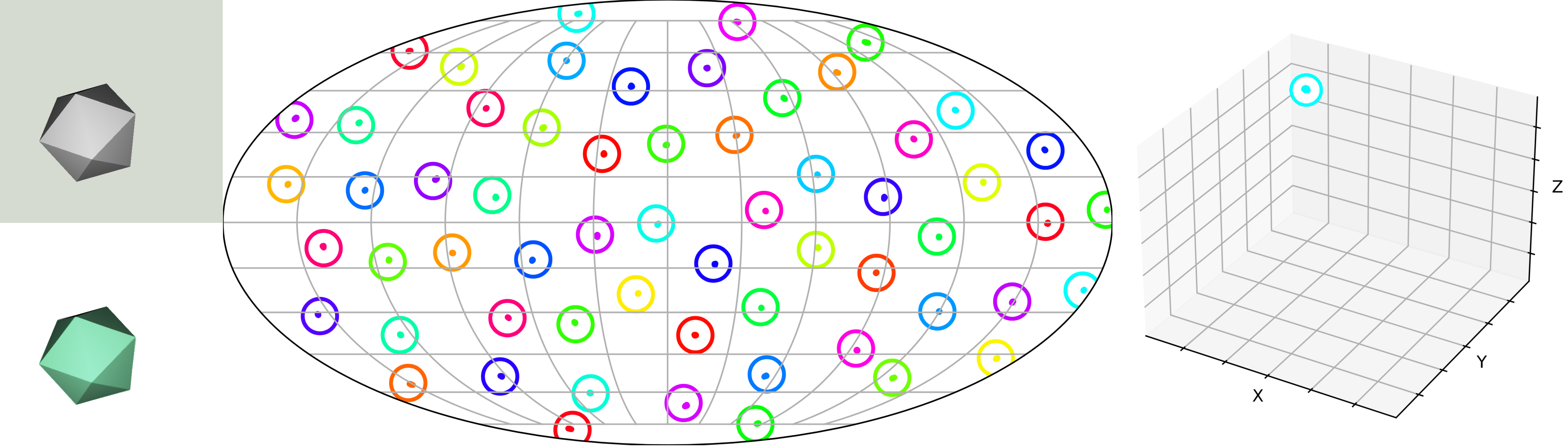}
    \includegraphics[width=.8\textwidth]{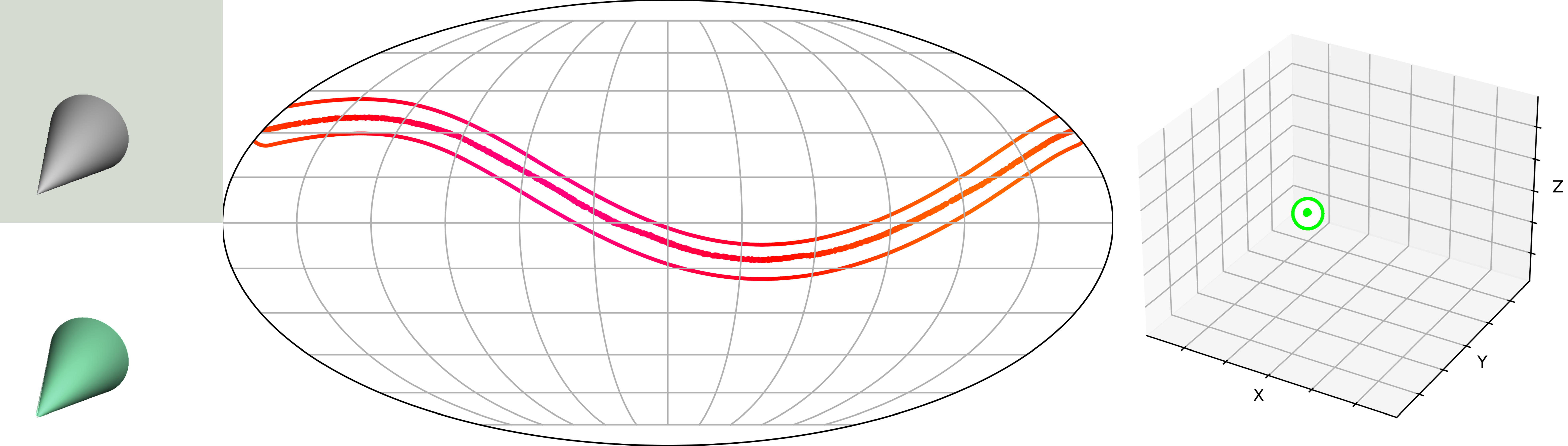}
    \includegraphics[width=.8\textwidth]{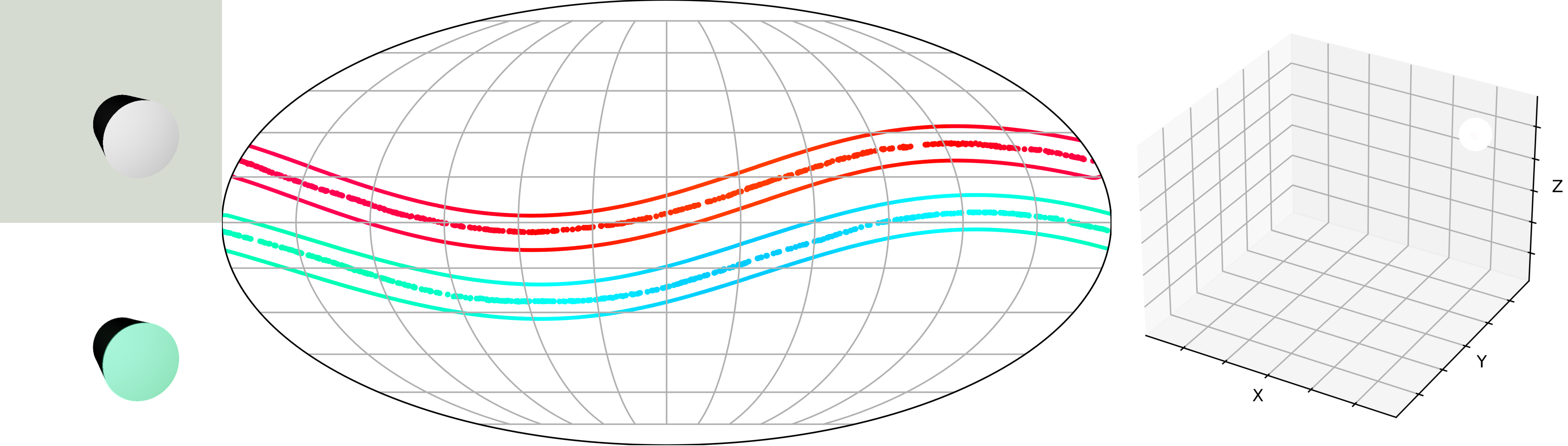}
    \caption{Visualization of our SYMSOL-T results. Please refer to Section~\ref{sec:visualization} for the detailed descriptions.}
    \label{fig:vis_symsolt}
\end{figure*}

In Fig.~\ref{fig:vis_symsolt}, we present the SYMSOL-T results obtained from our $SE(3)$ diffusion model for each shape. The model predictions are displayed in green and correlate to the corresponding original input images that are illustrated in gray. Our visualization strategy is described in Section~\ref{sec:experimental_setups}.
For each plot, we generate a total of $1,000$ random samples from our model. Please note that both the cone and the cylinder exhibit continuous symmetries. This causes the circles on $SO(3)$ to overlap densely and connect, which gives rise to tilde shapes on the sphere. In the case of $\mathbb{R}^3$, a single circle is present due to the unique solution for the translation. The samples generated from our score model are tightly concentrated in the center of each circle.  This evidence highlights the capability of our model to accurately capture equivalent object poses originating from either discrete or continuous symmetries.

\begin{figure*}[t]
  \centering
  \includegraphics[width=.9\linewidth]{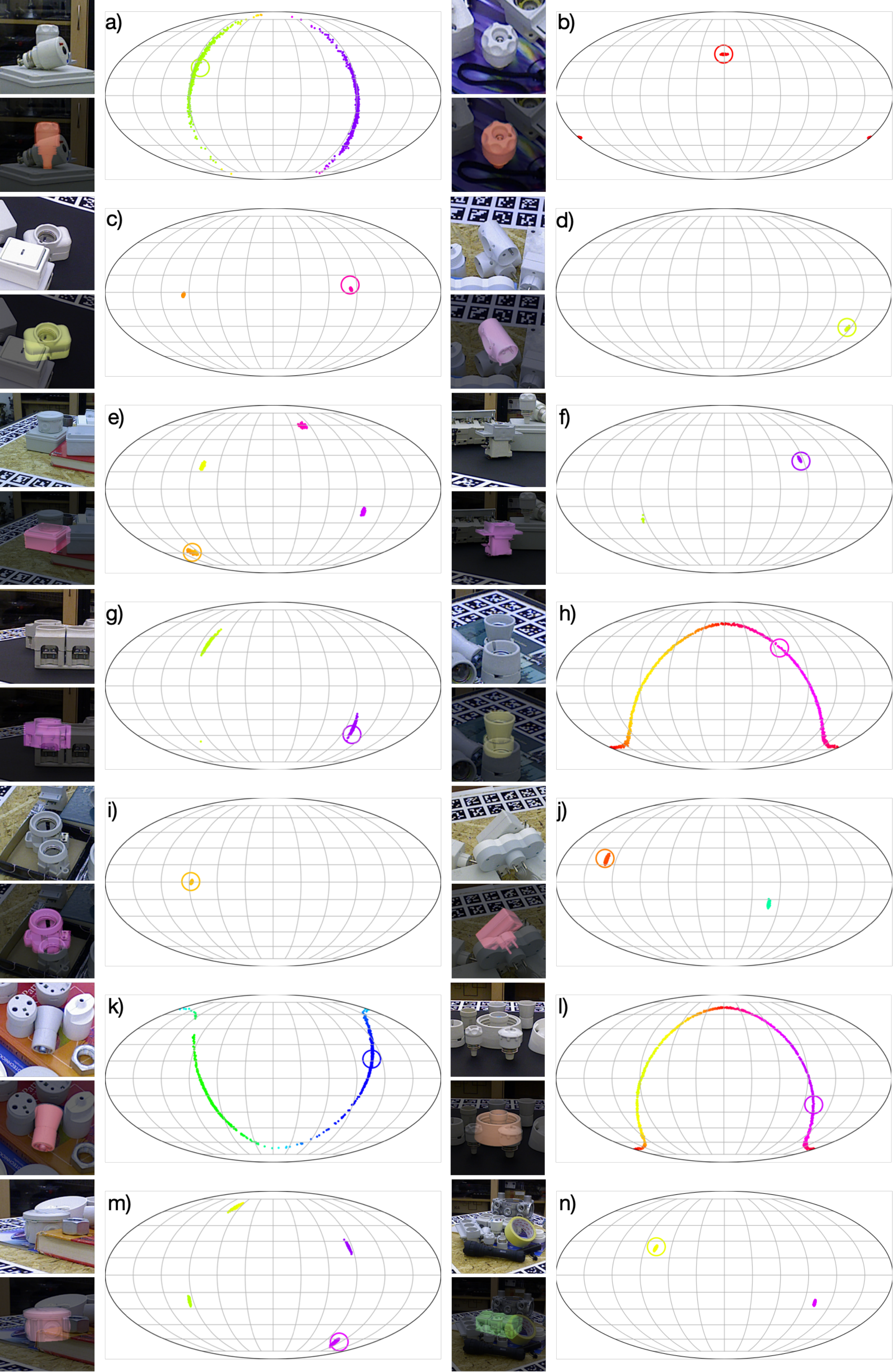}
  \caption{Visualization of our $SE(3)$ diffusion results on T-LESS.}
  \label{fig:tless-viz-sup}
  \vspace{-1em}
\end{figure*}

\section{Proofs}
\label{sec:proofs}

\subsection{Closed-Form of Stein Scores}

In this section, we present the derivation of the closed-form solution for the Stein scores. We begin with a revisitation of the Gaussian distribution on the Lie group $\mathcal{G}$, which is formulated as follows:
\begin{equation}
{\scriptsize
\begin{aligned}
    p_\Sigma(Y|X) &:= \mathcal{N}_\mathcal{G}(Y;X,\Sigma)\\
    &\triangleq \frac{1}{\zeta(\Sigma)}\text{exp}\left(-\frac{1}{2}\text{Log}(X^{-1}Y)^\top\Sigma^{-1}\text{Log}(X^{-1}Y)\right).
\end{aligned}
}
\end{equation}
To derive Eq.~(\ref{eq:score_form}), we utilize the definition of Stein scores, which is defined as the derivative of log-density of the data distribution with respect to the group element $Y\in\mathcal{G}$, expressed as follows:
\begin{equation}
{\scriptsize
\begin{aligned}
\nabla_Y &\log p_\Sigma(Y|X)^\top\\
    &= \frac{\partial}{\partial Y}\left(-\frac{1}{2}\text{Log}(X^{-1}Y)^\top\Sigma^{-1}\text{Log}(X^{-1}Y)\right)\\
    &= \frac{\partial}{\partial \text{Log}(X^{-1}Y)}\left(-\frac{1}{2}\text{Log}(X^{-1}Y)^\top\Sigma^{-1}\text{Log}(X^{-1}Y)\right)\frac{\partial \text{Log}(X^{-1}Y)}{\partial Y}\\
    &= - \text{Log}(X^{-1}Y)^\top\Sigma^{-1}\left(\frac{\partial \text{Log}(X^{-1}Y)}{\partial (X^{-1}Y)}\cdot\frac{\partial(X^{-1}Y)}{\partial Y}\right)\\
    &= - \text{Log}(X^{-1}Y)^\top\Sigma^{-1}\left(\mathbf{J}_r^{-1}(\text{Log}(X^{-1}Y))\cdot I\right)\\
    &= - \text{Log}(X^{-1}Y)^\top\Sigma^{-1}\mathbf{J}_r^{-1}(\text{Log}(X^{-1}Y)).
\end{aligned}
}
\end{equation}
Based on the above derivation, the closed-form solution for the Stein scores can be obtained as follows:
\begin{equation}
{\footnotesize
\begin{aligned}
    \nabla_Y \log p_\Sigma(Y|X) = -\mathbf{J}_r^{-\top}(\text{Log}(X^{-1}Y))\Sigma^{-1}\text{Log}(X^{-1}Y).
\end{aligned}
}
\end{equation}


\subsection{Left and Right Jacobians on $SO(3)$}
In this section, we present the derivation of Eq.~(\ref{eq:left_right_jacob}). Let $z=\left[z_x, z_y, z_z\right]\in\mathfrak{so}(3)$ and $\phi=\|z\|^2_2$. The skew-symmetric matrix induced by $z$ can therefore be represented as follows:
\begin{equation}
    z_\times = \left[\begin{matrix}
        0 & -z_z & z_y \\
        z_z & 0 & -z_x \\
        -z_y & z_x & 0 \\
    \end{matrix}\right]
\end{equation}
As demonstrated in~\cite{sola2018}, the left and the right Jacobian on $SO(3)$ can be expressed as the following closed-form expressions:
\begin{align}
\begin{aligned}
    \mathbf{J}_r(z) &= I - \frac{1-\cos\phi}{\phi^2}z_\times + \frac{\phi-\sin\phi}{\phi^3}z_\times^2\\
    \mathbf{J}^{-1}_r(z) &= I + \frac{1}{2}z_\times + \left(\frac{1}{\phi}-\frac{1+\cos\phi}{2\phi\sin\phi}\right)z^2_\times\\
    \mathbf{J}_l(z) &= I + \frac{1-\cos\phi}{\phi^2}z_\times + \frac{\phi-\sin\phi}{\phi^3}z_\times^2\\
    \mathbf{J}^{-1}_l(z) &= I - \frac{1}{2}z_\times + \left(\frac{1}{\phi}-\frac{1+\cos\phi}{2\phi\sin\phi}\right)z^2_\times.\\
\end{aligned}
\end{align}
As a result, Eq.~(\ref{eq:left_right_jacob}) of the main manuscript can be derived as follow:
\begin{equation}
    \mathbf{J}_l(z) = \mathbf{J}_r^\top(z),\qquad \mathbf{J}_l^{-1}(z) = \mathbf{J}_r^{-\top}(z).
\end{equation}


\subsection{Eigenvector of The Jacobians}

For the purpose of proving $\mathbf{J}_l(z)z=z$, we consider the derivative of exponential mapping on $\mathcal{G}$, where $k\in\mathbb{R}$ and $z\in\mathfrak{g}$. More specifically, by applying the chain rule on the derivative of the small perturbation $\text{Exp}(kz)$ on $\mathcal{G}$ with respect to $k$, we can obtain the resultant equation as follows:
\begin{equation}
    \frac{\partial \text{Exp}(kz)}{\partial k} = \frac{\partial \text{Exp}(kz)}{\partial(kz)}\frac{\partial (kz)}{\partial k} = \mathbf{J}_l(kz)z.
    \label{eq:chain_rule}
\end{equation}
On the other hand, by applying the differential rule, the following equations can be derived:
\begin{equation}
\begin{aligned}
\frac{\partial\text{Exp}(kz)}{\partial k}
    &= \lim_{h\to 0}\frac{\text{Log}(\text{Exp}((k+h)z)\text{Exp}(kz)^{-1})}{k}\\
    &= \lim_{h\to 0}\frac{\text{Log}(\text{Exp}(hz)\text{Exp}(kz)\text{Exp}(kz)^{-1})}{h} = z.
    \label{eq:differential_rule}
\end{aligned}
\end{equation}
By further combining Eqs.~(\ref{eq:chain_rule}) and~(\ref{eq:differential_rule}) and setting $k=1$, the following equation can be derived:
\begin{equation}
    \mathbf{J}_l(z)z = z.
    \label{eq:left_close}
\end{equation}
The resultant Eq.~(\ref{eq:left_close}) suggests that $z$ is an eigenvector of $\mathbf{J}_l(z)$. Please note that the same rule can also be employed to provide a proof for the right-Jacobian as follows:
\begin{equation}
    \mathbf{J}_r(z)z = z.
\end{equation}


\subsection{Closed-Form of Stein Scores on $SE(3)$}

In this section, we delve into the closed-form solution of Stein scores on $SE(3)$, which is referenced in Section~\ref{sec:methodology:simplify_score_se3}. Let $z=(\rho, \phi)\in\mathfrak{se}(3)$, where $\rho$ represents the translational part and $\phi$ denotes the rotational part. We define $\hat{\phi} = \|\phi\|^2_2$ and recall the inverse of the left-Jacobian on $SE(3)$ as follows:
\begin{equation}
    \mathbf{J}^{-1}_l(z) = \left[\begin{matrix}
        \mathbf{J}^{-1}_l(\phi) & \mathbf{Z}(\rho, \phi) \\
        0 & \mathbf{J}^{-1}_l(\phi)
    \end{matrix}\right],
\end{equation}
where $\mathbf{Z}(\rho, \phi)=-\mathbf{J}^{-1}_l(\phi)\mathbf{Q}(\rho, \phi)\mathbf{J}^{-1}_l(\phi)$. The complete form of $\mathbf{Q}(\rho, \phi)$ is defined in~\cite{sola2018, barfoot2014} as follows:
\begin{equation}
{\footnotesize
\begin{aligned}
\mathbf{Q}(&\rho, \phi) = \frac{1}{2}\rho_\times + \frac{\hat{\phi} - \sin\hat{\phi}}{\hat{\phi}^3}(\phi_\times\rho_\times + \rho_\times\phi_\times + \phi_\times\rho_\times\phi_\times)\\
    & -\frac{1-\frac{\hat{\phi}^2}{2}-\cos\hat{\phi}}{\hat{\phi}^4}(\phi^2_\times\rho_\times + \rho_\times\phi^2_\times -3\phi_\times\rho_\times\phi_\times)\\
    & - \frac{1}{2}\left(\frac{1-\frac{\hat{\phi}^2}{2}-\cos\hat{\phi}}{\hat{\phi}^4} - 3\frac{\hat{\phi}-\sin\hat{\phi}-\frac{\hat{\phi}^3}{6}}{\hat{\phi}^5}(\phi_\times\rho_\times\phi^2_\times + \phi^2_\times\rho_\times\phi_\times) \right).
\end{aligned}
}
\label{eq:q_def}
\end{equation}
From the Eq.~(\ref{eq:q_def}), an essential property can be observed and expressed as follows:
\begin{equation}
    \mathbf{Q}^\top(-\rho, -\phi) = \mathbf{Q}(\rho, \phi).
    \label{eq:q_eq}
\end{equation}
Based on the above derivation, the closed-form expression of the inverse transposed right-Jacobian on $SE(3)$ combined with the property outlined in Eq.~(\ref{eq:q_eq}) can be derived as follows:
\begin{equation}
\begin{aligned}
\mathbf{J}_r^{-\top}(z)
    &= \left(\mathbf{J}_l^{-1}(-z)\right)^\top\\
    &= \left[\begin{matrix}
        \mathbf{J}^{-1}_l(-\phi) & \mathbf{Z}(-\rho, -\phi)\\
        0 & \mathbf{J}^{-1}_l(-\phi)
    \end{matrix}\right]^\top\\
    &= \left[\begin{matrix}
        \mathbf{J}^{-1}_r(\phi) & -\mathbf{J}_r^{-1}(\phi)\mathbf{Q}(-\rho, -\phi)\mathbf{J}^{-1}_r(\phi)\\
        0 & \mathbf{J}^{-1}_r(\phi)
    \end{matrix}\right]^\top\\
    &= \left[\begin{matrix}
        \mathbf{J}^{-\top}_r(\phi) & 0\\
        -\mathbf{J}_r^{-\top}(\phi)\mathbf{Q}^\top(-\rho, -\phi)\mathbf{J}^{-\top}_r(\phi) & \mathbf{J}^{-\top}_r(\phi)
    \end{matrix}\right]\\
    &= \left[\begin{matrix}
        \mathbf{J}^{-1}_l(\phi) & 0\\
        -\mathbf{J}_l^{-1}(\phi)\mathbf{Q}(\rho, \phi)\mathbf{J}^{-1}_l(\phi) & \mathbf{J}^{-1}_l(\phi)
    \end{matrix}\right]\\
    &= \left[\begin{matrix}
        \mathbf{J}^{-1}_l(\phi) & 0\\
        \mathbf{Z}(\rho, \phi) & \mathbf{J}^{-1}_l(\phi).
    \end{matrix}\right]
\end{aligned}
\end{equation}

The closed-form solution of Stein score on $SE(3)$ can then be computed by the definition of Stein score as follows:
\begin{equation}
\nabla_Y \log p_\sigma(\tilde{X}|X) = -\frac{1}{\sigma^2}\left[\begin{matrix}
        \mathbf{J}^{-1}_l(\phi) & 0\\
        \mathbf{Z}(\rho, \phi) & \mathbf{J}^{-1}_l(\phi)
    \end{matrix}\right]z.
\end{equation}
After examining the derivation process, it is clear that this computation involves the costly calculation of Jacobians, and does not confer any computational benefits when using automatic differentiation. However, by adopting the surrogate score presented in Eq.~(\ref{eq:surrogate_score}), it is possible to reduce the computation of the Jacobian $\mathbf{J}^{-\top}_r(z)$, while simultaneously improving performance, as explained in Section~\ref{sec:surrogate_vs_autograd}.

\end{document}